\documentclass[journal]{IEEEtran}

\pdfoutput=1

\usepackage{graphicx,times,amsmath,algorithm,algorithmic}
\usepackage{bm}
\usepackage{colortbl}
\usepackage{subfig}

\usepackage[table,dvipsnames,prologue]{xcolor}
\definecolor{mylightgray}{gray}{0.9}

\usepackage[normalem]{ulem}  % for strike through \sout

\usepackage[bookmarks=true]{hyperref}
\hypersetup{
    colorlinks=true,       % false: boxed links; true: colored links
    linkcolor=black,       % color of internal links
    citecolor=black,       % color of links to bibliography
    filecolor=black,        % color of file links
    urlcolor=black,        % color of external links
    linktoc=page            % only page is linked
}

\usepackage{slashbox}
\usepackage{multirow}
\usepackage{array}
\usepackage{arydshln}

\newlength{\originalwidth}
\newcommand{\forcedoublecolumnfigurestart}{}
\newcommand{\forcedoublecolumnfigureend}{}
\newenvironment{fullfigure*}[1][tbp]
{\begin{figure*}[#1]\forcedoublecolumnfigurestart}
{\forcedoublecolumnfigureend\end{figure*}}

\usepackage{longfbox}

\makeatletter
\newdimen\@tempdimd
\makeatother

\newcommand{\tightframe}[1]{
	\lfbox[rounded,border-color=gray!90,border-width=0.033em,border-radius=0.3em]{#1}
}

\newcommand*{\vcenteredhbox}[1]{
	\begingroup
		\setbox0=\hbox{#1}\parbox{\wd0}{\box0}
	\endgroup
}

\newcommand{\pieScale}{0.75} 
\newcommand{\pcaScale}{0.25} 
\newcommand{\pieLegendScale}{0.85} 

\newcommand{\pcaLeftMargin}{-0.34cm}
\newcommand{\pcaRightMargin}{-0.57cm}

\newcommand{\pieLeftMargin}{-0.3cm}
\newcommand{\pieRightMargin}{-0.9cm}

\newcommand{\pieIdLeftMargin}{-0.48cm} 
\newcommand{\pieIdRightMargin}{-0.4cm}

\newcommand{\barPlotPath}{images-pdf/e28-64evo-fit}
\newcommand{\pcaPlotPath}{images-pdf/e28-64evo-pca1D}
\newcommand{\piePlotPath}{images-pdf/e28-agents-states}
\newcommand{\plotPath}{images-pdf}

\newcommand{\pcaPlot}[1]{
	\vcenteredhbox{
	  \hspace{\pcaLeftMargin}
	  \includegraphics[scale=\pcaScale]{\pcaPlotPath/#1}
	  \hspace{\pcaRightMargin}
	}
}
\newcommand{\piePlotCustomMargin}[3]{
	\vcenteredhbox{
		\hspace{#2}
		\includegraphics[scale=\pieScale]{\piePlotPath/#1}
		\hspace{#3}
	}
}
\newcommand{\piePlot}[1]{
	\piePlotCustomMargin{#1}{\pieLeftMargin}{\pieRightMargin}
}
\newcommand{\piePlotTableI}[1]{
	\begin{tabular}{cc}
	\hspace{\pieIdLeftMargin}  \tightframe{1}  \hspace{\pieIdRightMargin} & \piePlot{#1}
	\end{tabular}
}
\newcommand{\piePlotTableII}[2]{
	\begin{tabular}{cc}
	\hspace{\pieIdLeftMargin}  \tightframe{1}  \hspace{\pieIdRightMargin} & \piePlot{#1} \\
	\hspace{\pieIdLeftMargin}  \tightframe{2}  \hspace{\pieIdRightMargin} & \piePlot{#2}
	\end{tabular}
}

\newcommand{\piePlotTableV}[5]{
	\begin{tabular}{cc}
	\hspace{\pieIdLeftMargin}  \tightframe{1}  \hspace{\pieIdRightMargin} & \piePlot{#1} \\
	\hspace{\pieIdLeftMargin}  \tightframe{2}  \hspace{\pieIdRightMargin} & \piePlot{#2} \\
	\hspace{\pieIdLeftMargin}  \tightframe{3}  \hspace{\pieIdRightMargin} & \piePlot{#3} \\
	\hspace{\pieIdLeftMargin}  \tightframe{4}  \hspace{\pieIdRightMargin} & \piePlot{#4} \\
	\hspace{\pieIdLeftMargin}  \tightframe{5}  \hspace{\pieIdRightMargin} & \piePlot{#5}
	\end{tabular}
}
\newcommand{\pieOrderDescCell}[3]{
	\hspace{#2}
	\vcenteredhbox{ \resizebox{#1}{!}{
		\begin{math}
		\begin{array}{lllllll}
			\text{Pub. $\downarrow$} & \text{Pub. $\downarrow$} &
			\text{Pub. $\uparrow$} & \text{Pub. $\uparrow$} & & & \\
			\text{Priv. $\downarrow$} & \text{Priv. $\uparrow$} &
			\text{Priv. $\downarrow$} & \text{Priv. $\uparrow$} & &
			\text{Point $\downarrow$} & \text{Point $\uparrow$}
		\end{array}
		\end{math}
	}}
	\hspace{#3}
}
\newcommand{\pieOrderDescRow}{
	& &
	\hspace{0.07cm}$\longleftarrow$ Public \hfill Private $\longrightarrow$\hspace{-3.35cm} &
	\pieOrderDescCell{0.365\textwidth}{-0.1cm}{-0.75cm}
}
\newcommand{\pieLegendCell}[3]{
	\multicolumn{1}{l}{ \vcenteredhbox{
		\rule{0cm}{0.075cm}
		\hspace{#1}
		\resizebox{0.405\textwidth}{!}{
		        \begin{minipage}[t]{#3\textwidth}
		          \vspace{0pt}
			  \includegraphics*[scale=\pieLegendScale]{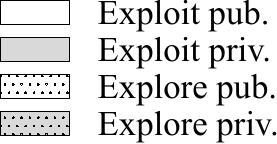}
		        \end{minipage}
		        \begin{minipage}[t]{#3\textwidth}
		          \vspace{0pt}
			  \includegraphics*[scale=\pieLegendScale]{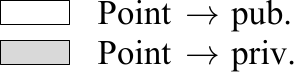}
		        \end{minipage}
		}
		\hspace{#2}
	}}
}
\newcommand{\pieLegendRow}{
	\multicolumn{1}{l}{} & \multicolumn{1}{l}{} & \multicolumn{1}{l}{} &
	\pieLegendCell{-0.15cm}{-1.4cm}{0.17}
}

\begin{document}

\title{Evolving Strategies for Competitive\\Multi-Agent Search}

\author{Erkin~Bah\c{c}eci, Riitta~Katila, and~Risto~Miikkulainen% <-this % stops a space

\thanks{This research was supported in part by NSF under grants SBE-0914796, IIS-0915038, and DBI-0939454.}
\thanks{Erkin Bah\c{c}eci and Risto Miikkulainen are with the
Department of Computer Science, University of Texas at Austin,
Austin, TX, 78712, USA (e-mail: {\tt erkin@cs.utexas.edu};
{\tt risto@cs.utexas.edu}).}
\thanks{Riitta Katila is with the
Department of Management Science \& Engineering, Stanford University,
Stanford, CA, 94305, USA (e-mail: {\tt rkatila@stanford.edu}).}
}

\maketitle

\thispagestyle{empty}

%%%%%%%%%%%%%%%%%%%%%%%%%%%%%%%%%%%%%%
%%%%%%%%%%%%%%%%%%%%%%%%%%%%%%%%%%%%%%

\begin{abstract}

While evolutionary computation is well suited for automatic discovery in engineering, it can also be used to gain insight into how humans and organizations could perform more effectively. Using a real-world problem of innovation search in organizations as the motivating example, this article first formalizes human creative problem solving as competitive multi-agent search (CMAS). CMAS is different from existing single-agent and team search problems in that the agents interact through knowledge of other agents' searches and through the dynamic changes in the search landscape that result from these searches. The main hypothesis is that evolutionary computation can be used to discover effective strategies for CMAS; this hypothesis is verified in a series of experiments on the $\emph{NK}$ model, i.e.\ partially correlated and tunably rugged fitness landscapes. Different specialized strategies are evolved for each different competitive environment, and also general strategies that perform well across environments. These strategies are more effective and more complex than hand-designed strategies and a strategy based on traditional tree search. Using a novel spherical visualization of such landscapes, insight is gained about how successful strategies work, e.g.\ by tracking positive changes in the landscape. The article thus provides a possible framework for studying various human creative activities as competitive multi-agent search in the future.

\end{abstract}


%%%%%%%%%%%%%%%%%%%%%%%%%%%%%%%%%%%%%%
%%%%%%%%%%%%%%%%%%%%%%%%%%%%%%%%%%%%%%

\section{Introduction}

\IEEEpubidadjcol

Evolutionary computation (EC) is often used as a method for optimizing
engineering design, be it physical or abstract constructions, or
descriptions of complex processes such as lisp code or neural
networks. However, EC can also be used to gain insight into how humans
perform various tasks, and how they could perform them better. This
article focuses on such a topic: It presents a model of human problem
solving in competitive domains, and demonstrates how evolutionary
computation can be used to discover effective behavior in such domains.

More specifically, the study makes three contributions. First, it
develops \emph{competitive multi-agent search} (CMAS) as a
formalization of human problem solving. CMAS was
originally developed to understand how high-tech companies search for
technological innovations, but the same formalization can potentially
be applied to computational modeling of scientific discovery,
engineering problem solving, and art and design.  In CMAS, multiple
agents search for the same peaks (i.e.\ innovations) on the common
fitness landscape.  They each try to find as many and as high peaks as
possible over a given amount of time, representing the cumulative
value of their innovations. While searching, they can choose to share
information about what they find, or keep such information private. Furthermore, the
landscape is dynamic in that the fitness of the points can increase or
decrease when multiple agents discover them, representing the dynamic
valuation of innovations in the real world.
% They search in a non-stationary environment because both the knowledge of opponents' searches and the fitness landscape itself changes as the search progresses. As a result, agents need to be reactive in order to respond to those changes. This requirement makes both the problem and the solution method different from standard search; the traditional formalizations of single-agent search do not adequately describe competitive multi-agent search.
The CMAS formalization is
useful because it makes it possible to characterize how humans solve
innovation and design problems, resulting in precise theories in
management science, psychology, and social science.  However, the
formalization also makes it possible to use such models to determine
how humans could perform better than they currently do, thus informing
both the individuals who are trying to solve these problems, and the
administrators that design policies to encourage innovation and
creativity.

\IEEEpubidadjcol

The second contribution, and the main one of this article, focuses on
this opportunity to do better. The article demonstrates that
\emph{evolutionary computation is a particularly good way to solve
CMAS problems}.  As an experimental platform, an abstract, general
CMAS domain is defined in terms of an $N\!K$ fitness landscape
\cite{kauffman:book93}.  A comprehensive array of basic search
strategies is created for this domain, based on local (exploitative)
search and long-range (exploratory) search using public and/or private
information about the landscape. An advanced search strategy is also
implemented based on tree search, representing a typical AI problem
solving method \cite{Korf:ai90}. These strategies are then used to
instantiate several different competitive environments, by including
competitors with different strategies.  New strategies are
evolved for each environment in order to perform better than the
existing ones.  The results show that (1) evolution can discover
customized strategies that perform well in each environment, (2) it
can discover general strategies that perform well across many
different environments, and (3) the good evolved strategies are more
complex than the basic intuitive strategies, employing different
strategies at different times, exhibiting optimal preference for acting
publicly or privately depending on the particular environment, and
resulting in overall principles such as riding a wave of dynamically
increasing landscape.

The third contribution is a technical one: a novel spherical
visualization for NK fitness landscapes. This visualization maintains
the continuity of the original high-dimensional landscape while
reducing it to an intuitive 3D surface. A focal point is selected, and
continuity is maintained by representing points further away with
lesser resolution. This visualization is useful in illustrating the
search strategies: For instance, it makes it strikingly clear why the
wave-riding behavior is so effective. However, the visualization is
general, and could be useful for any study involving high-dimensional
binary spaces.

The article thus shows that CMAS is a potentially useful way to study
problem solving in the real world, and that evolutionary computation
is an effective way to gain insight into such problems.

%%%%%%%%%%%%%%%%%%%%%%%%%%%%%%%%%%%%%%
%%%%%%%%%%%%%%%%%%%%%%%%%%%%%%%%%%%%%%

\section{Background}
\label{sec:related}

This study is motivated by the real-world problem of innovation search
in organizations. The formalization of this problem, CMAS, builds on
single-agent and team-search methods, but extends them with
competitive and cooperative dynamic interactions between agents. It
also builds on agent-based modeling, but applies that general approach
to doing innovation search in a landscape that changes dynamically
because of agent actions. The main contribution is to show that
evolutionary computation is a good way to discover effective solution
strategies for CMAS. While many different evolutionary approaches
could be used, the particular one tested in this article is based on
NEAT neuro-evolution \cite{stanley:ec02} as one representative
approach. With NEAT, strategies can be represented naturally with
neural networks whose complexity is matched with the task.

\subsection{Organizational Theory}
\label{sc:orgtheory}

Even though the main focus of this article is on evolutionary
optimization of CMAS strategies, it is useful to review the motivation
from the perspective of a real-world example of CMAS, that of
organizational theory in management science. The example serves to
make the issues and the motivation concrete.

Search (i.e.\ institutional problem solving) in the organizational
theory literature is typically thought to take place in a
\emph{knowledge space}, conceptualized as a landscape. In innovation
search, for example, firms generate, recombine, and manipulate
knowledge within a pool of technological possibilities
\cite{levinthal:jebo81}, and such activity can be tracked, for
example, by using patents (e.g.\ \cite{katila:amj02}). 
% For instance, the early gaming industry (Atari, Sega, Sony) searched primarily by exploration based on private knowledge: Each company create their own game console technology. In contrast more recently, companies such as Nintendo have exploited more public knowledge, utilizing existing patented technologies in graphics, GPU, and AI [********* citation?]
Such a search can be represented in an $N\!K$ landscape where $N$ corresponds to
dimensions of knowledge and $K$ determines how complex the relationships
between them are \cite{katila:nk10,Levinthal:ms1997,gavetti:asq00}. This
work has led to several insights. One is that firms that search more
frequently and further away from their current knowledge bases
(i.e.\ \emph{explore}) are more likely to succeed
\cite{greve:amj03,katila:amj02}.  Another is that firms typically
search in exactly the opposite way: too little and too close
(i.e.\ \emph{exploit}), and therefore need to find effective
strategies to resist such local tendencies \cite{helfat:ms94}.

Despite these insights, the focus of organizational search research
has been relatively narrow. Prior efforts typically assessed a firm's
innovation activities only relative to its own behavior, i.e.\ as
\emph{single-agent search}. Only recently, researchers have started to
conceptualize search beyond its single-agent roots and to incorporate
competition. There is emerging research on situations in which firms
learn from their competitors and on why such learning is sometimes
difficult \cite{greve:asq00,rivkin:ms00}, as well as research on how
competitors interact dynamically \cite{katila:asq08}. However, to date
these studies have been conceptual and statistical only; competition
has not been integrated in any formal models of organizational search.

The first contribution of this article is to do so, i.e.\ to create a
formalization that can be used to study such competitive processes
with computational techniques.  It contributes thus to management
science, but it also contributes to AI, by defining a new and
interesting class of search problems relevant to the real world.
Although the motivation comes from the specific example of
organizational search, the same formalization should be useful in
understanding a range of problem solving activities in human
societies, such as scientific discovery, innovation in engineering,
and art and design. % >>>> partial repetition? (see introduction)

The second, main contribution of this article is then to show that new
computational techniques are useful in this new domain. The next
subsection reviews traditional search techniques in artificial
intelligence and evolutionary computation, pointing out why they
are not a good fit with CMAS problems.

\subsection{Search Algorithms}

Traditionally, research on search algorithms has focused on two types of search methods: search performed by a single agent and search performed by a team of similar cooperating agents. Single-agent search methods, such as A* \cite{Hart:ssc68} and iterative-deepening A* (IDA*) \cite{korf:dfid},
have been used in well-defined search domains, including path finding and scheduling problems. Such methods are understood well theoretically and guarantee optimal solutions to a problem, but they are impractical to utilize if the search space of the problem is too large.

On the other hand, with team-search methods the individual team members search for peaks in a fitness landscape in parallel; every point in the search space has a certain height corresponding to its fitness value, and the knowledge of all agents is collected into a single pool.
These search methods are appropriate in problems that are inherently large or not well-defined, such as antenna design \cite{lohn2004evolutionary} and robot control \cite{valsalam:ieeetec07}.
The theory of these methods is less well developed and they do not usually guarantee optimal solutions. 

Team-search methods are typically inspired by various types of biological and natural systems, such as evolution \cite{mitchell:gaintro96}, swarm behavior \cite{james2001swarm}, water drops \cite{Hosseini2009}, and gravitational particle interactions \cite{Rashedi2009}. For instance, Particle Swarm Optimization \cite{kennedy1995particle} and Ant Colony Optimization \cite{Dorigo2004} are inspired by social swarm behaviors in nature: They make use of multiple agents that represent solutions to optimization problems.

Competition has been incorporated into single-agent search methods by extending them to two-player games \cite{pearl:heuristics} and multi-agent adversarial games \cite{zuckerman:tciaig2011}. The search for the best move for a player proceeds by first considering all moves of that player and then a subset of the moves of the opponents (utilizing techniques such as alpha-beta pruning \cite{knuth:ai1975,korf:ai1991}).
However, since such search methods rely on enumerating all possible moves of at least one player, they are not practical with a large number of moves.

Competitive elements exist in team search as well. For instance, in evolutionary search, population members compete to propagate their genes, although the population as a whole cooperates to produce a single good solution. Inter-population competition has been incorporated into evolutionary search in a coevolutionary arrangement, where different populations try to outdo each other in the task \cite{pollack:alife96,Juille:sab96,werfel:ec00,stanley:jair04}. However, there is no absolute fitness; a team is considered successful simply if it does better than the other team. Moreover, the teams do not alter the fitness landscape and therefore do not influence others' search.

Therefore, neither single-agent nor team-based search is a good fit
with CMAS problems. In CMAS, interactions among agents and between the
agents and the environment need to be taken into account explicitly.
Agent-based modeling provides a framework for doing that, as will be
described next.

\subsection{Agent-Based Modeling}

Agent-based modeling has been used extensively in various fields including search and optimization in computer science \cite{Dorigo2004,kennedy1995particle,Knight1993}, as well as real-world social and economical interactions in political science \cite{axelrod1997complexity} and economics \cite{Epstein1999,Holland1991}. The idea is to model each agent explicitly, with the goal that global patterns emerge from this process. The agent-based approach is therefore an appropriate formulation for competitive multi-agent search.

In the multi-agent systems literature, there are many examples of
domains where the agents compete to solve problems
\cite{hoen:lamas06}.  CMAS can be seen as a special case of such
problems, characterized by four special properties: (1) Competitive
multi-agent search is modeled as search for the same highest peaks in
a common landscape. (2) The agents may not necessarily know about the
other agents' searches, and may or may not inform them about their own
searches. (3) Their search actions have an effect on the landscape
that is visible to all agents. (4) The agents' search strategies are
stochastic, representing the bounded rationality of real-world agents
such as human decision makers and organizations \cite{Simon1958}.

While some of these properties have been addressed in prior research,
together they define a new and interesting problem class.  It shares
with prior work the idea of agents and their interactions as the
appropriate level of modeling. However, building on these specific
properties, it may be possible to develop a specific formalization and
approach for CMAS problems that makes it easier to understand and
solve such problems. This is the goal of this article.

The main hypothesis tested is that while it is possible to formulate
search strategies by hand, and adapt traditional single-agent search
strategies to CMAS, better strategies can be discovered automatically
by evolutionary optimization. In order to verify this hypothesis, a
particular approach is developed using evolution of neural networks
with the NEAT method \cite{stanley:ec02}. While other methods are
possible as well, representing CMAS strategies as pattern-producing
neural networks (such as CPPNs \cite{stanley:gpem07} or CPGs
\cite{chiel:jcn99}) is a potentially powerful approach, as will be described
in detail in Section~\ref{sec:evolving}.  While such networks could be
evolved through different methods, NEAT has been previously applied to
them extensively, and will therefore be used as the default
implementation in this article as well.

\subsection{Neuro-Evolution of Augmenting Topologies (NEAT)}
\label{sec:neat}

NEAT (Neuro-Evolution of Augmenting Topologies; \cite{stanley:ec02}) is a method for evolving neural networks that adjusts both the topology and the weights as part of the learning process (for other such methods, see \cite{harvey:ras1997,Hutt:annga2003,moriguchi:gecco2012}). It is based on three synergetic ideas (see e.g.\ \cite{stanley:ec02} for details):

First, the initial population of neural networks consists of minimally
connected individuals with no hidden nodes (i.e.\ nodes other than
input and output nodes). The networks gradually become more complex
through mutations that add nodes and connections. Only those additions
to the topology that improve performance are kept, which helps find
small solutions to the problem. Starting with minimal topology also
speeds up learning since the number of connection weights to be
optimized during evolution, i.e.\ the size of the search space for
connection weights, is minimal \cite{stanley:ec02}.

Second, crossover between individuals with different topologies is made possible by keeping an innovation number for each gene in the genome of a neural network. Innovation numbers are used to match genes that have similar historical origin. They are an abstraction of homology in biological evolution, which is the mechanism for aligning similar genes during crossover \cite{sigal:jmb72}. Keeping these numbers for each gene circumvents the expensive task of matching topologies of networks for crossover (see \cite{stanley:ec02} for details on innovation numbers).

The third component of NEAT is that innovation in population members is protected by separating the population into species depending on similarity. When a structural mutation alters an individual considerably, the individual may not initially perform as well as other population members. If this happens, that individual will not survive even though this mutation might have led to a better-performing individual after some optimization. Speciation protects such individuals by putting networks that are too different from others into separate species, allowing them to be optimized within the species first. To prevent the whole population from being reduced to a single species, explicit fitness sharing \cite{goldberg:icga87} is used. This principle means that the fitness within the species is shared among its members, dividing the fitness of each individual by the size of its species, preventing species from becoming too large.

NEAT has been shown to be successful in several open-ended design
domains, such as vehicle control and collision warning
\cite{kohl:gecco06} and controlling video game agents
\cite{stanley:ieeetec05}. Most importantly, it has been particularly
effective in evolving Compositional Pattern Producing Networks
(CPPNs), i.e.\ networks that produce spatial patterns
\cite{stanley:gpem07}. The approach developed in this article will make
use of this idea: Agent strategies will be encoded as 4D and 2D
patterns, as will be described in Section~\ref{sec:evolving}.

%%%%%%%%%%%%%%%%%%%%%%%%%%%%%%%%%%%%%%
%%%%%%%%%%%%%%%%%%%%%%%%%%%%%%%%%%%%%%

\section{Approach}
\label{sec:method}

In CMAS, multiple agents search for the highest peaks in the same landscape simultaneously. The agents either share or hide their knowledge about the landscape from other agents, and their searches change the landscape dynamically. They select search actions (either exploit locally or explore globally) based on a strategy that is encoded as CPPNs, and evolved through the NEAT method. This section describes the landscapes, the agent models, the memory types and search actions they use, how the strategies are represented and generated from CPPNs, and how multiple agents are simulated.

\subsection{Fitness Landscape}

As the search space for the experiment, abstract $N\!K$ fitness
landscapes \cite{kauffman:book93} are used. $N\!K$ landscapes are
$N$-dimensional hypercubes that assign fitness values to the points of
the space such that the \emph{ruggedness} of the landscape can be
adjusted (using the $K$ parameter). Such landscapes have been used
extensively to model human problem solving, such as innovation search
\cite{Levinthal:ms1997,Anderson:os1999,Gavetti:asq2000}. Compared to
alternatives such as game theoretical models, the characteristics of
the environment can be readily incorporated into an $N\!K$ simulation,
a large number of agents can be included, and they can be boundedly
rational, making it easier to draw relevant insights \cite{Levinthal:ms1997}.

The $K$ parameter specifies the level of interaction among the $N$ dimensions and can take the values between 0 and $N-1$. When $K$ is 0, the fitness landscape is single-peaked and smooth. One can go from the lowest-fitness point to the single peak by simply following the fitness gradient (i.e. separately flipping each bit that causes the fitness to increase). With small $K$, the fitness landscape becomes a little rugged, but highest peaks are concentrated in a region. When $K$ is $N-1$, the landscape becomes fully random with many peaks distributed all over the space.

\begin{figure}
\centering
    \includegraphics[width=\columnwidth]{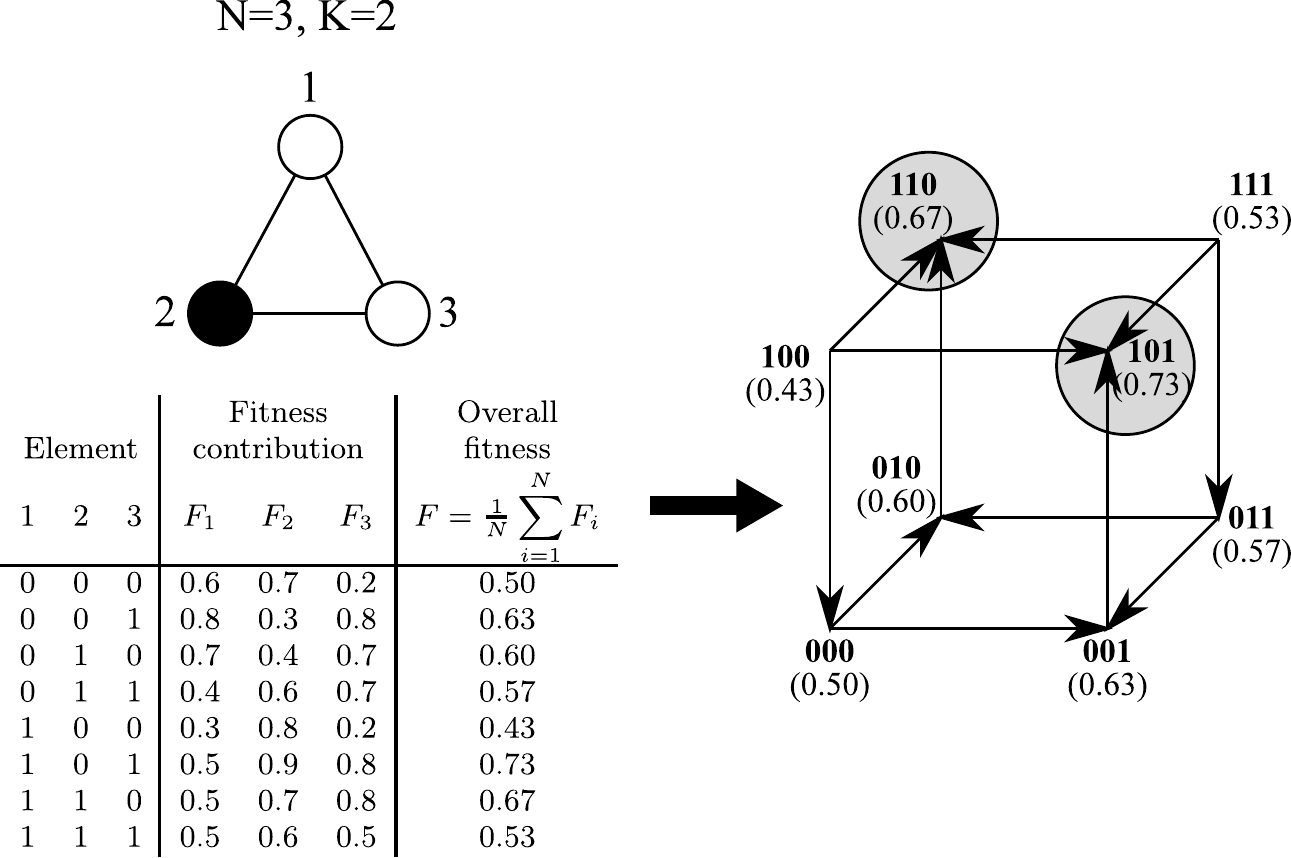}
    \caption{An $N\!K$ fitness landscape, where $N=3$ is the number of dimensions or bits of each point and $K=2$ is the number of other dimensions that interact with each dimension. For example, dimension 2 interacts with dimensions 1 and 3 (i.e.\ the two bits that follow it). To obtain the fitness of a point, e.g., 010, the fitness contribution values for its element values are averaged (i.e.\ the third row in the table).
The arrows on the hypercube represent the direction of increasing fitness. Also shown on the hypercube are the peaks in the $N\!K$ fitness landscape, indicated with shaded circles. The fitness values of all points constitute the complete fitness landscape. $N\!K$ landscapes are useful because they are general and the difficulty (i.e.\ ruggedness) can be adjusted. Such abstract landscapes can be used as a platform to study search methods.}
    \label{fig:NK}
\end{figure}

More specifically, each point in the search space is encoded as a bit string of length $N$. The fitness of a point is calculated by taking the average of the fitness contributions of each bit in the bit string for that point. Each bit's contribution depends on the value of $K+1$ bits: that bit and the $K$ bits that interact with that bit. As suggested by Kauffman \cite{kauffman:book93}, for bit $i$ the interacting bits are the $K$ bits that follow it, i.e., bits $i+1$, $i+2$, ..., $i+K$ (mod~$N$). These $K+1$ bit values are used as a key to look up fitness-contribution values from a table, generated randomly from a uniform distribution. This table has $N$ random values for each $(K+1)$-bit key, of which there are $2^{K+1}$.
Figure~\ref{fig:NK} shows how the fitness is calculated for an $N\!K$ fitness landscape with $N = 3$ and $K=2$. In this case, the table consists of $3\times8$ fitness contribution values.

Dynamic fitness landscapes are modeled through \emph{flocking}. That is, whenever an agent visits a point, the fitness of that point and those nearby change, depending on the \emph{flocking intensity} and \emph{flocking radius} parameters. The fitness of the area around a point defined by the flocking radius is multiplied by the flocking intensity. Two types of flocking are used. With \emph{boosting}, flocking intensity is greater than 1.0 and the region rises, whereas with \emph{crowding}, flocking intensity is less than 1.0 and the region sinks. There is no limit on the number of times a point can be visited. Therefore, when used in isolation, boosting and crowding will cause a point's fitness value to approach 1.0 and 0.0, respectively, with every visit to that point. These changes make it possible to model the dynamics of fitness landscapes in applications such as innovation search, where boosting corresponds to expanding demand in new markets (such as tablet computers) and crowding to the saturation of existing markets (such as desktops) \cite{katila:sej2012}. The details of the simulated agents are described next.

\subsection{Agents}

Each search agent is a software entity that looks for high-fitness points in the given fitness landscape. The behavior of an agent depends on the current state of the fitness landscape, the agent's strategy, and the current points in its memory.

Formally, the search agent's knowledge ${\cal X}(t)$ of the landscape and its topography at time $t$ consists of the points ${\bf x}_i$ with fitness values $z({\bf x}_i)$ ($1\leq i \leq t$), where $z$ is the fitness function:
\begin{equation}
\label{eq:eq1}
  {\cal X}(t) = \{[{\bf x}_1, z({\bf x}_1)], [{\bf x}_2, z({\bf x}_2)], ... [{\bf x}_t, z({\bf x}_t)]\}.
\end{equation}

The agent moves to the next (i.e.\ $(t+1)$\textsuperscript{th}) point using a search strategy $S$ based on what the agent already knows about the landscape (i.e. points visited by that agent and other agents):
\vspace*{-0.5ex}
\begin{equation}
\label{eq:eq2}
  {\bf x}_{t+1} = S[{\cal X}(t)].
\vspace*{-0.5ex}
\end{equation}

An agent's strategy $S$ consists of two components that determine how the agent will use its current knowledge. The first strategy component, $S_{1}$, specifies which type of memory (i.e.\ public or private, Section~\ref{sec:memory}) and which search method (i.e.\ explore globally or exploit locally, Section~\ref{sec:search-methods}) the agent will employ. The second strategy component, $S_{2}$, specifies in which type of memory the agent will place the last point it found.

\begin{algorithm}[t]
\caption{Simulation algorithm}
\label{alg:simulation}
\begin{algorithmic}[1]
%\STATE Read simulation parameters
\STATE Initialize simulation and agents.
\FOR {each time step until the maximum number of steps is reached}
    \FOR {each agent $a_i$}
        \STATE Advance $a_{i}$ one time step (Algorithm~\ref{alg:agent}).
        \STATE Record any landscape visits and public memory updates by $a_{i}$. \label{alg:line:record}
    \ENDFOR
    %\STATE Advance simulation state one time step
    \STATE Apply recorded landscape visits by updating fitness landscape. \label{alg:line:apply-visits}
    \STATE Apply recorded public memory updates. \label{alg:line:apply-memory-updates}
    %\PRINT average and best performance among all agents at current time step
\ENDFOR
\end{algorithmic}
\end{algorithm}

The simulation runs in discrete time steps, where every time step each agent is allowed to move according to its strategy and based on its current knowledge (Algorithm~\ref{alg:agent}).
At the beginning of a time step, each agent probabilistically selects a search method and a source memory for the search starting point. This point is given as input to the search method chosen by the agent to find a new high-fitness point.
At the end of each time step, if the agent has discovered a point that is better than the previous one, it schedules that point to be placed into the destination memory.

Even though the simulation advances the agents sequentially (Algorithm~\ref{alg:simulation}), the collective outcome of each time step in the simulation is independent of the agent execution order, due to the delayed execution of the side effects of agent actions. That is, every time step the simulator records landscape visits performed and public memory updates scheduled by agents, without actually making any changes immediately (line~\ref{alg:line:record} in the algorithm). The recorded agent visits are performed at the end of each time step after all the agents complete their search steps, and the resulting landscape changes are carried out (line~\ref{alg:line:apply-visits}). Similarly, public memory updates scheduled during a time step are applied at the end of that step (line~\ref{alg:line:apply-memory-updates}). The simulation is then carried out until the agents have searched a significant part of the space (100 points in the experiments in this article). The memory types will be described next.

\subsection{Memory}
\label{sec:memory}

Agents can place points in two types of memory: public and private. In terms of the innovation search example, public memory corresponds to public knowledge through patents, and private memory to trade secrets. Public memory can be accessed by all agents and it serves as a common knowledge base among agents. That is, through the use of public memory, agents can communicate the discovery of high-fitness points to the other agents. This information in turn attracts the other agents to those good points, potentially benefiting all agents. However, use of public memory also makes it likely for all agents to spend their time in the same region of the search space, and through dynamic landscape changes, may lead to decreasing fitness.

On the other hand, private memory is unique to each agent. Each agent can place points in its own private memory, where they are hidden from other agents. When agents make use of their private memory, they are more likely to spread out and explore different regions of the fitness landscape, leading to increased coverage of the search space.

Because memory contents change at each time step, public and private memory are denoted as ${\cal X}_{\mathrm{pub}}(t)$ and ${\cal X}_{\mathrm{priv}}(t)$, respectively. Before performing each search action, an agent takes the best point currently in the memory (either public or private, depending on the strategy), and performs a search using this point as the starting location. The next section describes the two search methods.

\begin{algorithm}[t]
\caption{Agent's algorithm to complete one time step}
\label{alg:agent}
\begin{algorithmic}[1]
\STATE Pick a search method and source memory probabilistically using $S_{1}$ strategy.
\STATE Perform one search step starting with the best point in the source memory.
\IF {found a better point than the last one}
    \STATE Pick a destination memory probabilistically using $S_{2}$ strategy.
    \STATE Schedule placement of the new point in the destination memory.
\ENDIF
\end{algorithmic}
\end{algorithm}

\subsection{Search Methods}
\label{sec:search-methods}

The agents employ two search methods probabilistically depending on their $S_{1}$ strategy: the exploit search method, i.e.\ taking a local step, and the explore search method, i.e.\ making a long jump in the search space. These two methods are motivated by how agents, such as innovating firms, search in the real world (Section~\ref{sc:orgtheory}; \cite{March:os1991}; note that thus these terms in this article do not refer to deterministic and stochastic actions like they do in the reinforcement learning literature, but instead to the length of the search step as they do in the management science literature).

The \emph{exploit} search method starts with a given point in the search space. It then tries to discover new high-fitness points that are immediate neighbors of that point, i.e.\ are at 1-bit distance from it. Each new point is generated by flipping one bit in the point's $N$-bit representation. If the new point has a better fitness than the starting point, that point is placed in memory. Otherwise, the search continues by flipping another bit of the starting point, and so on until all bits are tried. The order of the flipped bits is a random permutation of numbers 1 through $N$.

The \emph{explore} search method also starts with a given point, but it obtains new points in a different way. From the starting point, it jumps to a new point that is \emph{not} an immediate neighbor. More specifically, it generates a new point by flipping multiple random bits of the starting point simultaneously and continues to do so until the new point has higher fitness than the starting point, or the maximum number of jump attempts is reached. The number of bits to be flipped is also chosen randomly within a range given in simulation parameters. For the experiments in this article this range is set to [0.5, 1.0], which means that a new point in an exploration step is obtained by flipping between 50-100\% of bits of the starting point. In this manner, the exploration action is a relatively long jump, and therefore distinctly different from the exploitation action.

These search methods provide two actions for agents to perform on the starting point. They are selected stochastically based on the agent's strategy, as described next.

\begin{table}[t]
\begin{center}
{\setlength{\tabcolsep}{0.3em}
\resizebox{\columnwidth}{!}{
\begin{tabular}{|l|c:c:c:c|}
\hline
\multirow{2}{*}{\backslashbox{State:}{Action:}} &
Exploit with & Exploit with & Explore with & Explore with \\
 & public mem. & private mem. & public mem. & private mem. \\
\hline
Public: low fit. & \multirow{2}{*}{0.1} & \multirow{2}{*}{0.2} & \multirow{2}{*}{0.3} & \multirow{2}{*}{0.4} \\
Private: low fit. &&&& \\
\hline
Public: low fit. & \multirow{2}{*}{0.0} & \multirow{2}{*}{1.0} & \multirow{2}{*}{0} & \multirow{2}{*}{0} \\
Private: high fit. &&&& \\
\hline
Public: high fit. & \multirow{2}{*}{0.005} & \multirow{2}{*}{0.995} & \multirow{2}{*}{0} & \multirow{2}{*}{0} \\
Private: low fit. &&&& \\
\hline
Public: high fit. & \multirow{2}{*}{0} & \multirow{2}{*}{0} & \multirow{2}{*}{0.9} & \multirow{2}{*}{0.1} \\
Private: high fit. &&&& \\
\hline

\end{tabular}
}
}
\end{center}
\caption{An example $S_1$ strategy component, where each row represents a state, and each column represents an action. $S_1$ consists of $4 \times 4$ probability values, one per action-state combination, which are used for selecting an action (i.e.\ a search method and a starting point) given the state (i.e.\ the binary-valued fitness of the best points in public and private memory). Each row of probability values adds up to 1.0, and determines what the agent will do in the corresponding state. For instance, the second row of this particular strategy specifies that when the best public point has low fitness and best private point has high fitness, the agent will exploit that public point with 0.1 probability, exploit that private point with 0.2 probability, explore starting with that public point with 0.3 probability, or explore starting with that private point with 0.4 probability.}
\label{tbl:S1}
\end{table}

\subsection{Agent Strategy}

Agents select among the two search methods using a search strategy. To represent this strategy, the state of each agent is converted to a discrete form based on whether the considered points' fitness values are less than 0.5 (i.e.\ \emph{low fitness}) or more than 0.5 (i.e.\ \emph{high fitness}).
The two components of an agent's strategy, $S_1$ and $S_2$, consist of a set of probability values for each discrete state of the agent.
Tables~\ref{tbl:S1} and \ref{tbl:S2} show example $S_{1}$ and $S_{2}$ strategies, respectively. Each row represents a discrete state of an agent, and each column represents an action. The agents choose their actions probabilistically depending on their current state, according to the probabilities in these tables.

\begin{figure}
\begin{center}
\resizebox{\columnwidth}{!}{
\begin{tabular}{|l|}

\multicolumn{1}{l}{
    \pieOrderDescCell{0.36\textwidth}{-0.5cm}{-0.6cm}
} \\  \hline

\piePlotCustomMargin{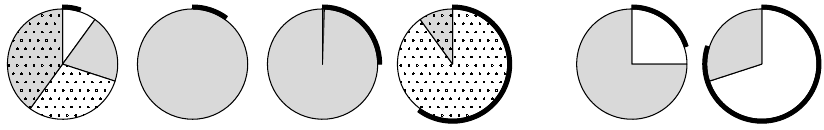}{-0.36cm}{-0.55cm} \\  \hline

\pieLegendCell{0cm}{-2.6cm}{0.22} \\

\end{tabular}
}
\end{center}
\vspace{-0.3cm}
\caption{
A pie-chart depiction of the example agent strategy shown in Tables~\ref{tbl:S1} and \ref{tbl:S2}. Each of the circles corresponds to a row of one of those two tables (i.e. a state): four circles on the left for $S_{1}$ and two on the right for $S_{2}$, where $\downarrow$ and $\uparrow$ represent low and high fitness, respectively. Each shading pattern corresponds to a column of one of the two tables (i.e. an action). The size of each slice represents the probability of the action in the corresponding table row and column. This chart format allows visualization of the 20 probability values of a strategy in a compact way.
Additionally, the black band around each circle indicates the average percentage of time the agent spent in the state that corresponds to the circle (i.e. a row in Tables~\ref{tbl:S1} and \ref{tbl:S2}).
The example percentages shown from left to right are 5\%, 10\%, 25\%, and 60\% for $S_{1}$, and 20\% and 80\% for $S_{2}$. The total area of gray slices overall indicates how much private memory is used, whereas the amount of dotted shading in the first four circles tells the ratio of exploration. For instance, the first circle shows a row with non-zero probabilities for all actions, whereas the second circle specifies that only one action is possible in the corresponding state (i.e. with 1.0 probability). In certain cases, the probabilities for all but one action are close but not equal to 1.0 (e.g. third row in Table~\ref{tbl:S1}), which is seen in the visualization as a sliver (e.g. the third circle). Such small but nonzero probabilities were often discovered in the evolutionary experiments (Tables~\ref{tbl:evolved-strategies-sparse} and \ref{tbl:evolved-strategies-dense}), and they turned out to make a significant difference in performance compared to similar fixed strategies where those probabilities are 0.0 (Figure~\ref{fig:evalComparisonEvo}).
}
\label{fig:strategy-pie-format}
\end{figure}

\begin{table}[t]
\begin{center}
\resizebox{\columnwidth}{!}{
\begin{tabular}{|l|c:c|}
\hline
\multirow{2}{*}{\backslashbox{State:}{Action:}} & Place in & Place in \\
 & public memory &  private memory \\
\hline
New point: low fitness & 0.25 & 0.75 \\
\hline
New point: high fitness & 0.7 & 0.3 \\
\hline

\end{tabular}
}
\end{center}
\caption{An example $S_2$ strategy component. $S_2$ consists of $2 \times 2$ probability values (for two actions and two states) for determining to which memory to place the new point given the discrete fitness of that point. Each row of probabilities adds up to 1.0, and determines what the agent will do in the corresponding state. In this example, the agent will e.g. place each new low-fitness point into public memory with 0.25 probability and into private memory with 0.75 probability. The agent strategies can be visualized graphically as shown in Figure~\ref{fig:strategy-pie-format}.}
\label{tbl:S2}
\end{table}

The $S_1$ strategy component is employed by agents to visit a new point. Using $S_1$, an agent selects both a search method and a starting point, i.e.\ the memory's best point from which the search begins, depending on the discrete-valued fitness of the best points of public and private memory. The state consists of the discrete-valued fitness of the two memories and the action is a combination of the search method and the starting point chosen. Thus, there are four possible states (low or high fitness for public memory's best point and low or high fitness for private memory's best point) and four possible actions (exploit or explore with public or private memory). The search yields a new point for the agent to visit.
Thus, $S_{1}$ can be formalized as
\begin{eqnarray}
\label{eq:S1}
{\bf x}_{t+1} &=& S_1({\cal X}_{\rm pub}(t), {\cal X}_{\rm priv}(t)).
\end{eqnarray}

Using the $S_2$ strategy component, agents determine where to put a newly discovered point (i.e.\ the action) depending on the fitness of the new point ${\bf x}_{t+1}$ (i.e.\ the state).
Thus, there are two states (low fitness or high fitness) and two actions (placing the point in public or private memory). Formally, $S_{2}$ is used to update knowledge, i.e.
\begin{eqnarray}
\label{eq:S2}
\begin{split}
{\cal X}(t+1) &= \{{\cal X}_{\rm pub}(t+1), {\cal X}_{\rm priv}(t+1)\}\\
                      &= S_2(z({\bf x}_{t+1}), {\cal X}_{\rm pub}(t), {\cal X}_{\rm priv}(t)).
\end{split}
\end{eqnarray}
where $z$ is the fitness function.

The $S_{1}$ and $S_{2}$ strategies determine how agents behave and how knowledge gets updated in the simulation. They can be represented directly as vectors of real numbers such as those in Tables~\ref{tbl:S1} and \ref{tbl:S2}, and visualized graphically in Figure~\ref{fig:strategy-pie-format}, and this representation is sufficient for the hand-coded strategies in this article. However, preliminary experiments on learning strategies showed that representing them as continuous patterns through CPPNs leads to better results. How CPPNs are used to encode agent strategies will be described next.

\begin{figure}[t]
\centerline{\includegraphics[width=0.4\columnwidth]{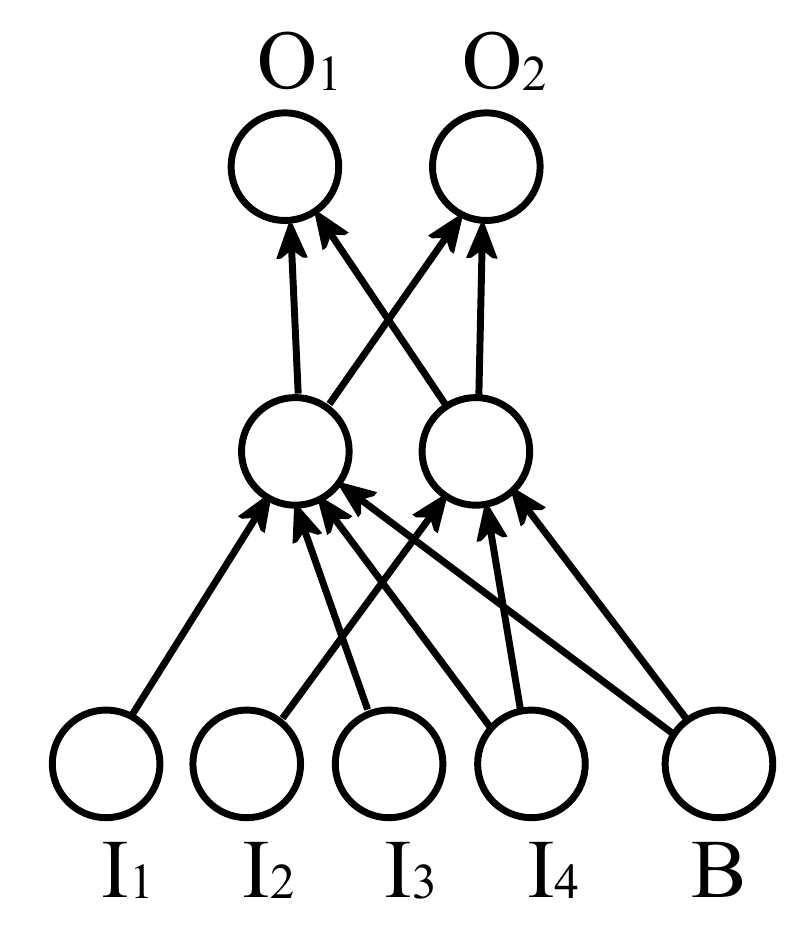}}
\caption{An example CPPN with four inputs, a bias input, and two outputs. The number of links and hidden nodes as well as weights of existing links are not fixed, and can change during evolution. Combinations of -1 and 1 values supplied to input nodes $I_1, I_2, I_3, I_4$ act as indices into the rows and columns of Tables~\ref{tbl:S1} and \ref{tbl:S2} (with -1 value representing index 0). For $S_1$, $I_1$ specifies whether public memory has low or high fitness, $I_2$ specifies whether private memory has low or high fitness, $I_3$ specifies whether the action is exploit or explore, and $I_4$ specifies whether the action is carried out using public or private memory; the output value of $O_1$ for each combination of the four inputs determines the action probability for the state and action specified by that combination, which is the value shown in the corresponding cell of Table~\ref{tbl:S1}.
Similarly, for $S_2$, $I_1$ specifies whether the new point has low or high fitness, and $I_4$ specifies whether the action is to place that point in public or private memory; the output value of $O_2$ for each combination of those two inputs determines the action probability for the state and action specified by that combination, which is the value shown in the corresponding cell of Table~\ref{tbl:S2}. While calculating probabilities for $S_{2}$, $I_2$ and $I_3$ are set to 0.0.
In this manner, strategies can be represented continuously as CPPNs, with $S_1$ and $S_2$ sharing the same network structure.
}
\label{fig:CPPN}
\end{figure}

\subsection{Encoding Strategy Patterns} %Generation of Strategies from CPPNs
\label{sec:evolving}

Compositional Pattern Producing Networks (CPPNs) are neural networks with nodes that have various activation functions. They are particularly useful for creating geometric patterns such as symmetric ones \cite{stanley:gpem07}.
Preliminary experiments on learning strategies showed that representing them as continuous patterns through CPPNs leads to better results than evolving a set of distinct probability values that correspond to strategy tables.

To represent strategies as CPPNs, note that the $S_1$ and $S_2$ strategy components can be seen as functions that output the probability values in Tables~\ref{tbl:S1} and \ref{tbl:S2}.
For instance in the case of $S_2$, the function takes two binary inputs: whether the new point has low or high fitness, and whether the destination memory being considered is the public or private memory. Each combination of these binary inputs generates one of the probability values in Table~\ref{tbl:S2}. This function can be implemented as a CPPN with two inputs and one output, which can be evolved with a neuro-evolution algorithm. 

On the other hand, in the case of $S_1$ there are four discrete inputs: whether the best point in public memory has low or high fitness, whether the best point in private memory has low or high fitness, whether exploit or explore action is being considered, and whether the source memory for that action is public or private memory. This function can also be represented as a CPPN with one output, but with four inputs instead of only two, to produce probability values as in Table~\ref{tbl:S1}.

To simplify the approach further, the functions for $S_1$ and $S_2$ can be represented together by a single CPPN with four inputs and two outputs (one output per strategy component; Figure~\ref{fig:CPPN}) instead of by two separate CPPNs with one output each. Utilizing such a combined CPPN for both $S_1$ and $S_2$ allows a single population of networks to be evolved. The components that belong together to evolve together, sharing common structure.

To generate the 16 probability values in $S_1$ (as in Table~\ref{tbl:S1}), all four inputs of the CPPN network are used. The inputs are set to each combination of -1 or 1 in turn, representing the different cells in the strategy table (where -1 corresponds to the 0 index). The activation of the first output unit is passed through a Gaussian or sigmoid function to limit its value within the (0, 1) range, and the resulting number entered into the corresponding $S_1$ table cell. After all cells have been filled in this manner, the rows of $S_1$ are normalized to add up to 1. If all values in a row are very small, then all cells in that row are set to equal probability. The process for $S_2$ is similar to $S_1$, except that only the first two inputs of the CPPN are used, and the second output unit of the network is used to calculate the values for the $S_2$ cells. This approach allows agent strategies to be evolved conveniently as CPPNs.

%%%%%%%%%%%%%%%%%%%%%%%%%%%%%%%%%%%%%%
%%%%%%%%%%%%%%%%%%%%%%%%%%%%%%%%%%%%%%

\section{Experiments}
\label{sec:exp}

This section provides an experimental analysis of CMAS using the simulation setup described in the previous section. The experiments aim to characterize various environments, determine what kind of strategies work best, and test the hypothesis that evolutionary optimization can be used to learn better agent strategies (in particular specific strategies for a given environment and general strategies that perform well across multiple environments). The agent strategies are encoded as CPPNs, and evolved using NEAT.
%and find out whether better general and specific strategies can be learned.

\subsection{Simulation Environments}
\label{sec:simEnv}

The simulation environments all had eight agents. The first agent's strategy was evaluated and the remaining seven agents were given the role of \emph{opponents}, all with a common manually specified and fixed strategy. The $S_1$ and $S_2$ components of the opponents' strategies were set to constant probabilities, resulting in a CMAS environment with specific characteristics. Six distinct opponent strategies (i.e. six different environments) were implemented: always exploit or always explore, and always using the best point from the public memory, always using the best point from the private memory, or using either of those two points with a 50\% probability. These hand-coded strategies were selected because they are intuitive, clear, and represent common strategies in innovation search \cite{katila:amj02,greve:amj03,helfat:ms94}.

The environments also varied in search space dimensions ($N$), resulting in two different densities of agents (i.e. number of agents divided by search space size) since the number of agents was the same: (1) sparse environments with $N=20$ and (2) dense ones with $N=10$. The $K$ parameter of the $N\!K$ landscapes was set to three in all environments to provide a moderate level of interaction among the $N$ dimensions. Such moderate interaction is motivated from the organizational search perspective \cite{knudsen:os07}. The fitness landscape changed via initial boosting and subsequent crowding as agents moved in the landscape, achieved by decaying the flocking intensity from 1.05 to 0.9 linearly for each point over 10 agent visits to the same point, with a flocking radius constant at 2.0. Due to this dynamic nature of the fitness landscape, which is directly related to how many agents there are on average per search space point, the environments are more naturally characterized by agent density rather than search space size. These environment parameter values were chosen because they resulted in meaningful innovation search behavior in preliminary simulations.

\begin{figure*}
\centering
\subfloat[Sparse environments]{
\includegraphics[width=0.9\textwidth]{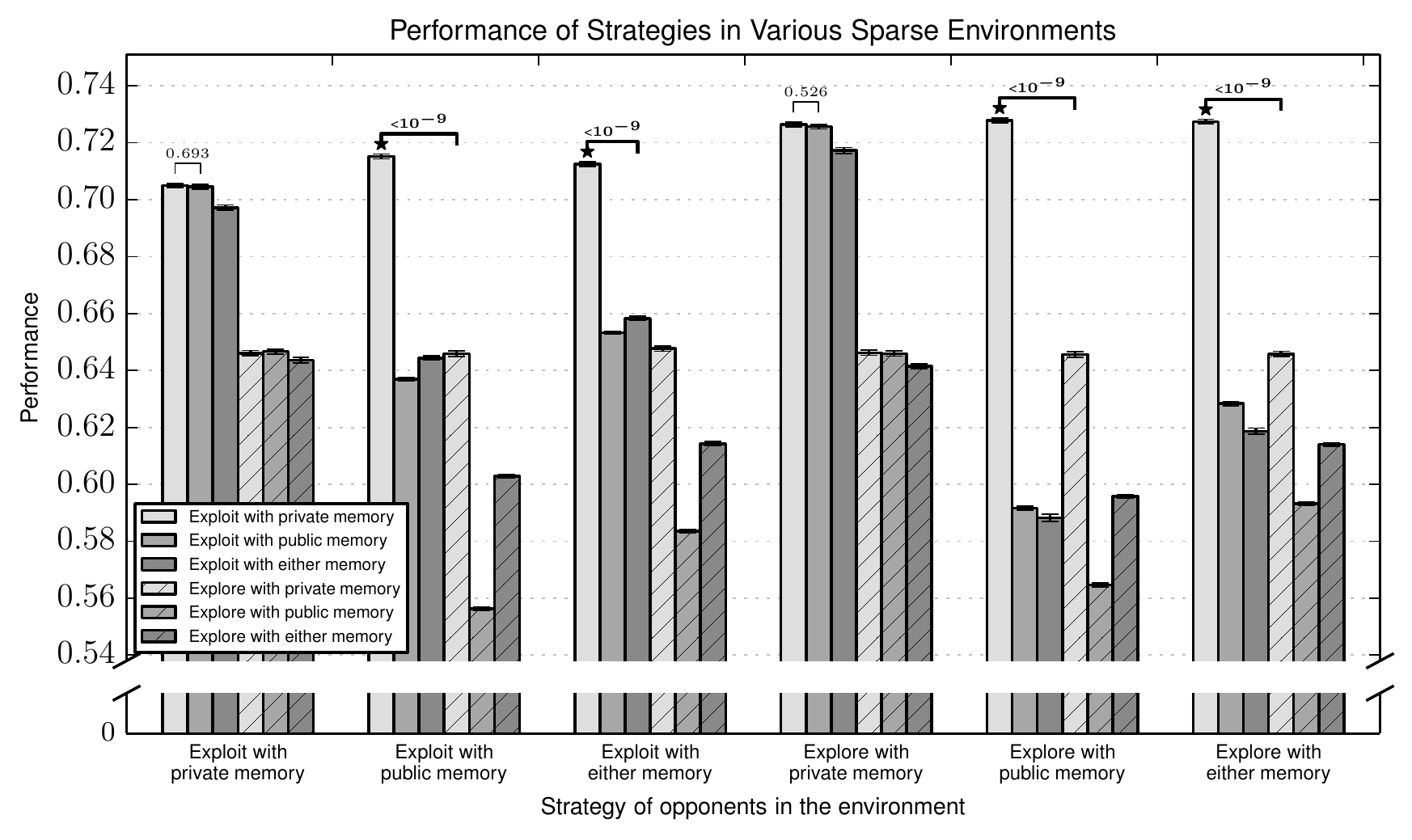}
\label{fig:evalComparisonManualSparse}
}\\
\subfloat[Dense environments]{
\includegraphics[width=0.9\textwidth]{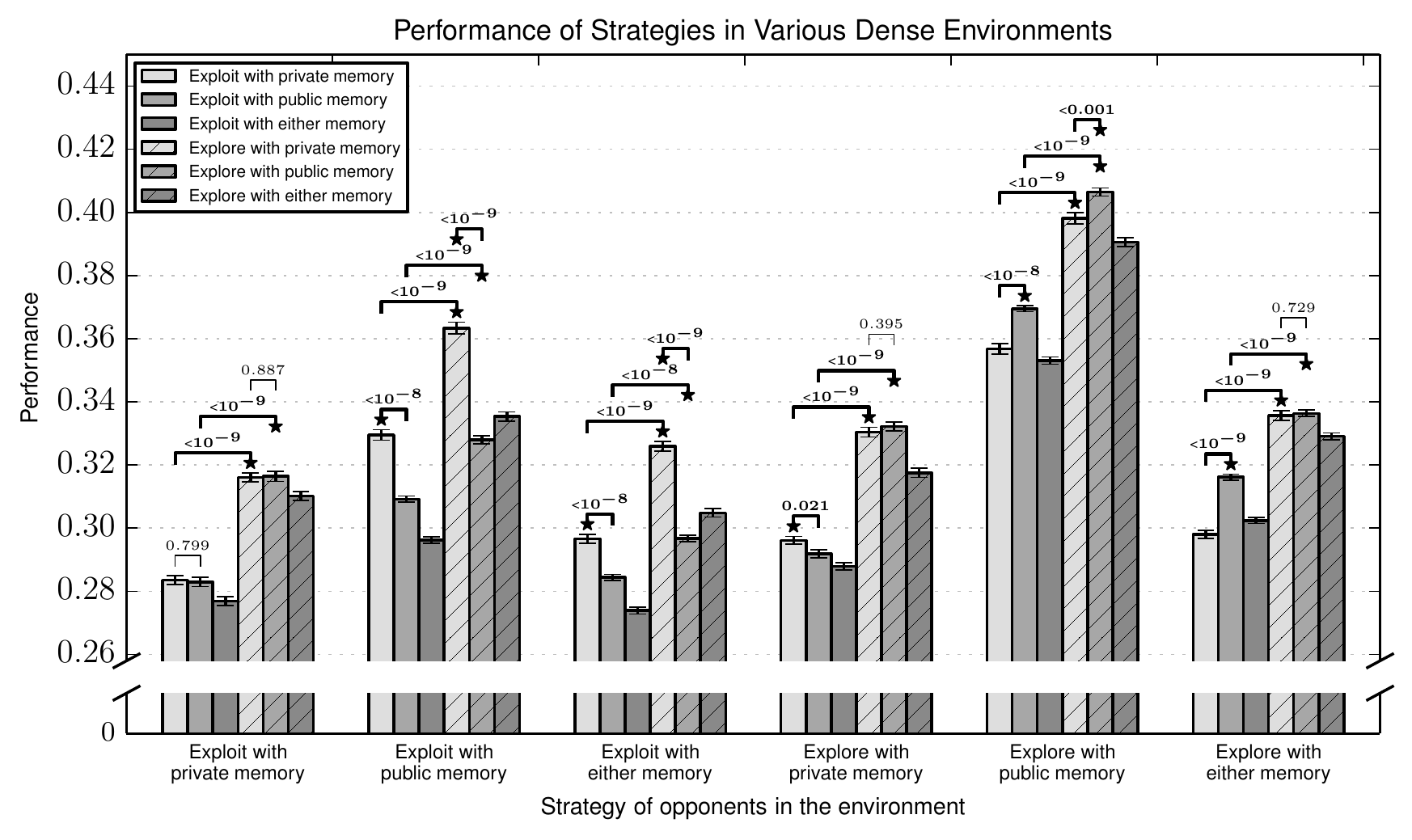}
\label{fig:evalComparisonManualDense}
}
\caption[]{Performance comparison among manually specified strategies in sparse ($N=20$) and dense ($N=10$) environments. Statistical significance is estimated between averages over 200 starting locations and shown for the fixed strategies in each environment.
Significant differences are indicated by stars.
Error bars denote one unit of standard error of the mean.
Note that because there is more interaction in the dense environment and therefore more crowding, the $y$-axes have different scales.
Each strategy (represented by a bar of different color) was evaluated in six different homogeneous environments with seven identical opponents whose strategy is identified along the $x$-axis. Performance on each environment was averaged over 200 evaluation runs. Different strategies perform best in different environments; finding an effective strategy for a given environment is possible and an important and interesting challenge.}
\label{fig:evalComparisonManual}
\end{figure*}

Performance with several distinct strategies in such environments is compared in Figure~\ref{fig:evalComparisonManual}. In each environment and for each evaluated strategy the simulation was repeated 200 times with randomized agent starting points in each run. The performance of the strategy was then calculated by averaging the performance across those runs, which in turn was defined as the average fitness of the points that the agent with that strategy visited during the 100 time steps (i.e the duration of the run). Since fitness of points in the search space varies between 0 and 1, so does each strategy's performance.

In the sparse environments (Figure~\ref{fig:evalComparisonManualSparse}), the \emph{exploit with private memory} strategy performed the best among all evaluated strategies (with $p$-value $< 10^{-9}$ compared to the one with the second highest mean performance), except in the environments where the opponents used only private memory. In those environments it tied with the \emph{exploit with public memory} strategy. Indeed, when the opponents never access the public memory, that memory becomes equivalent to the private memory for the evaluated agent. Thus, the successful strategy in the sparse environments can be described as \emph{exploit with a memory that is not accessible by the opponents}.

On the other hand, when the environment was dense (Figure~\ref{fig:evalComparisonManualDense}) exploring performed better than exploiting in general. However, the best memory type to use while exploring depended on the opponent strategy. For instance, when the opponents \emph{explored with public memory}, it was better for the evaluated agent to use public memory (with $p$-value $< 10^{-3}$), whereas when the opponents \emph{exploited with public memory}, it was better to use private memory (with $p$-value $< 10^{-8}$). Indeed, in this case the regions around good public memory points would generally become crowded and therefore have low fitness, making private memory search more effective.

These results suggest that different environments require different search strategies. Further, finding an effective strategy for a given environment is possible and constitutes an interesting and important challenge.

Obviously, such a hand-coded comparison can ever only include a small subset of all possible strategies. The ones included are prototypical and cover the space well, but there is no reason to believe that they are the best for the given environments.
In order to determine experimentally whether better strategies exist, a machine discovery method can be employed, as will be done in the next two sections. First, a strategy will be optimized for each environment separately; then, the same method will be used to create a general strategy by optimizing the average performance of a strategy across multiple environments.

\subsection{Evolving Strategies for a Particular Environment}
\label{sec:expEvolveParticular}

\begin{figure}[t]
\centerline{\includegraphics[width=\columnwidth]{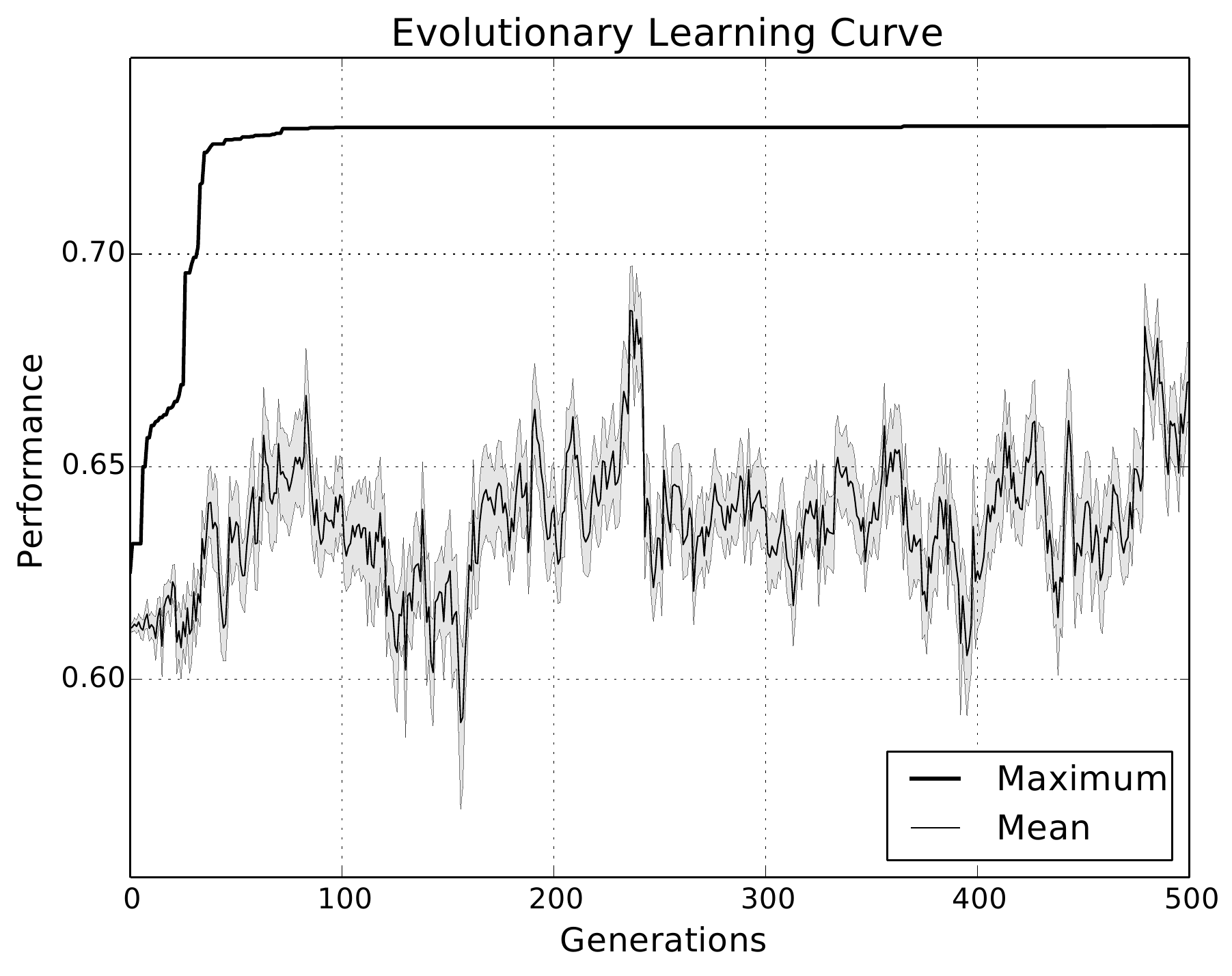}}
\caption{Learning curve for one of the evolutionary runs, showing the maximum and mean performance of the evolved population. The shaded region indicates standard error of the mean. Maximum performance usually plateaued well before 500 generations.}
\label{fig:learning-curve}
\end{figure}

To test the hypothesis that better strategies exist, strategies were evolved using NEAT with a population of size 100 in the same homogeneous environments as before. NEAT was allowed to run for 500 generations, which was the point by when performance plateaued in preliminary runs (Figure~\ref{fig:learning-curve}). Evolutionary runs were repeated 64 times. Other parameters were standard for NEAT \cite{stanley:ec02} or otherwise listed in Table~\ref{tbl:parameters} in the Appendix.

At each generation of an evolutionary run, the fitness of each population individual, i.e.\ each CPPN, was calculated by first generating the $S_1$ and $S_2$ probability tables from the CPPN, and then using those tables to control the single evaluated agent in the environment.
Sizes of evolved CPPNs vary, but average 26.6 and 26.9 nodes, and 75.4 and 75.6 links for sparse and dense environments, respectively.

\begin{table*}[tp]
\centering
\resizebox{\textwidth}{!}{
\begin{tabular}{|c|l|l|l|l|}
\hline
\multicolumn{2}{|l|}{\textbf{Sparse Environment}} &
\multirow{2}{*}{\textbf{PCA for All Evolved Strategies}} &
\multirow{2}{*}{\textbf{Sample Strategies}} \\
\cline{1-2}
\textbf{Id} & \textbf{Opponents} & & \\
\hline\hline

\pieOrderDescRow \\  \hline

1 & \textbf{Exploit} with \textbf{public} memory \hspace{-0.4cm} &
\pcaPlot{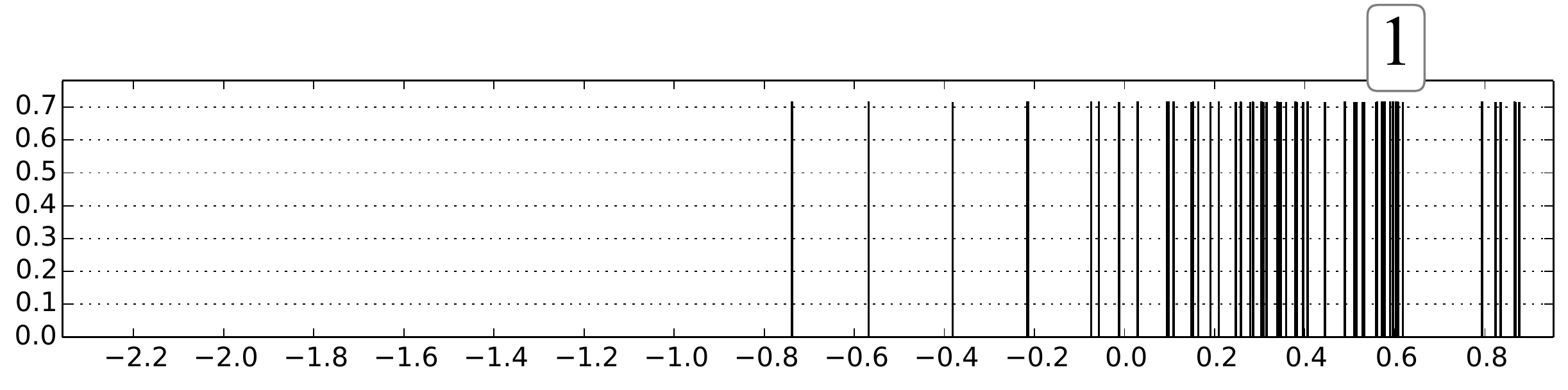} & \piePlotTableI{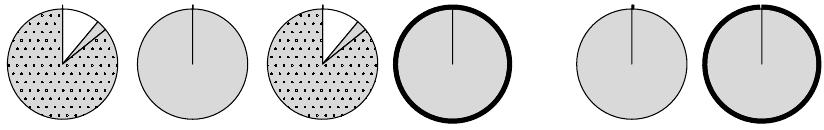} \\
\hline

2 & \textbf{Explore} with \textbf{public} memory \hspace{-0.4cm} &
\pcaPlot{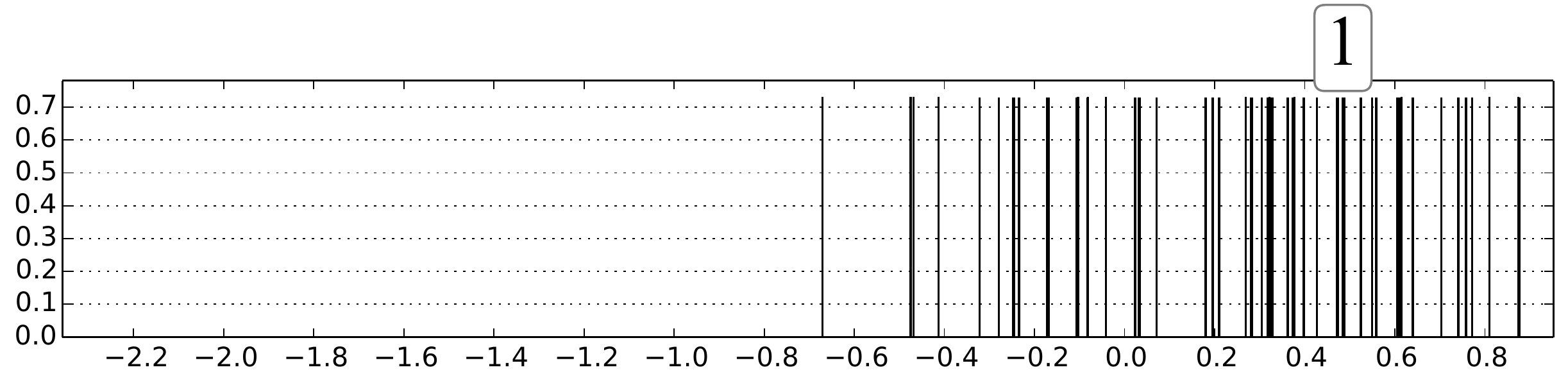} & \piePlotTableI{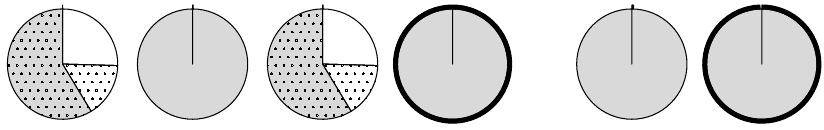} \\
\hline

3 & \textbf{Exploit} with \textbf{private} memory \hspace{-0.4cm} &
\pcaPlot{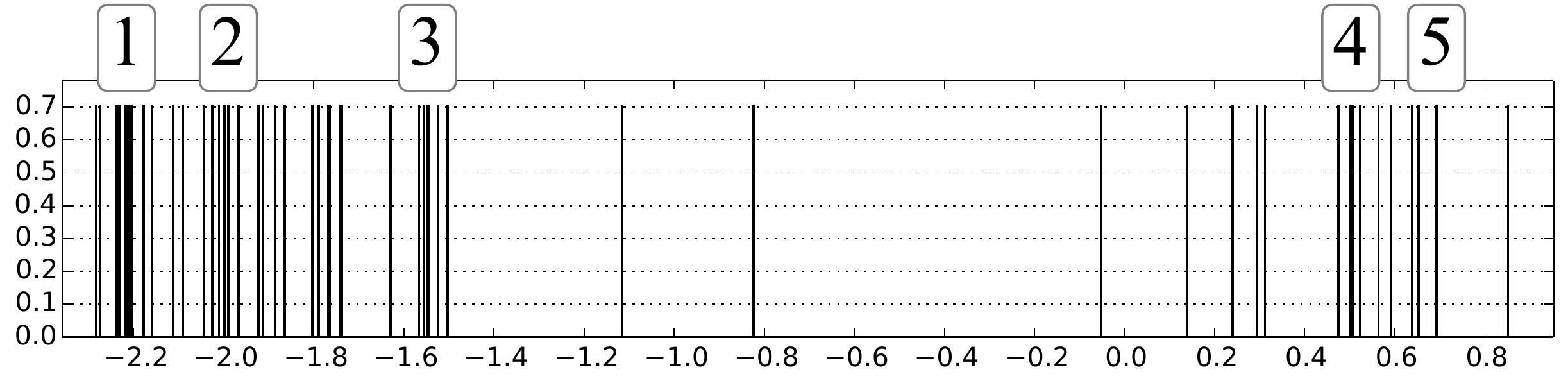} &
\piePlotTableV{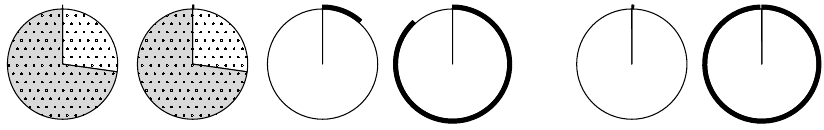}{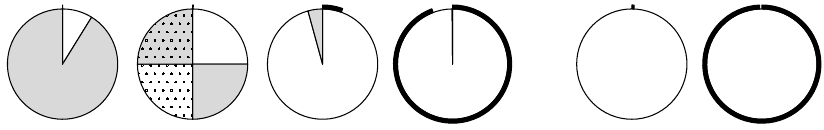}{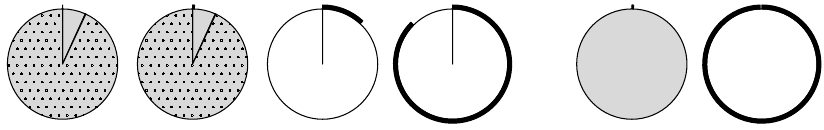}{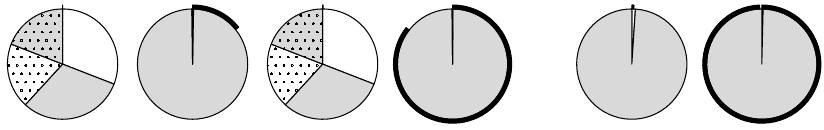}{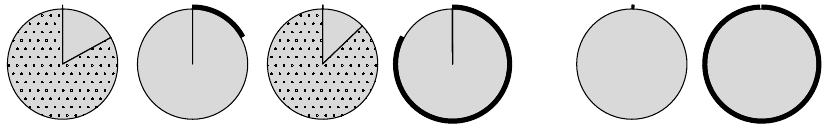} \\
\hline

4 & \textbf{Explore} with \textbf{private} memory \hspace{-0.4cm} &
\pcaPlot{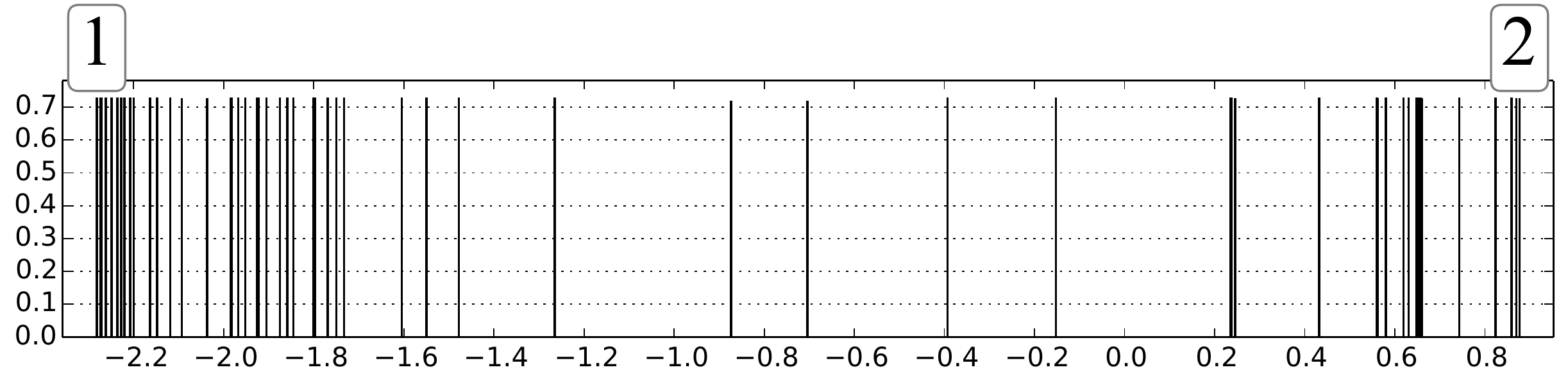} &
\piePlotTableII{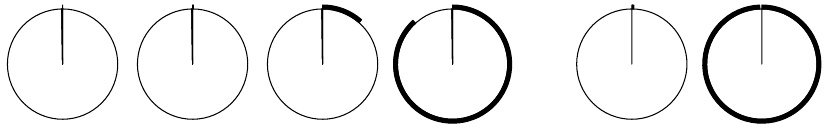}{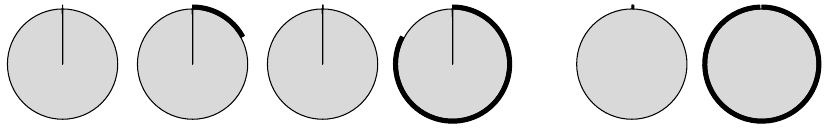} \\
\hline

5 & \textbf{Exploit} with \textbf{either} memory \hspace{-0.4cm} &
\pcaPlot{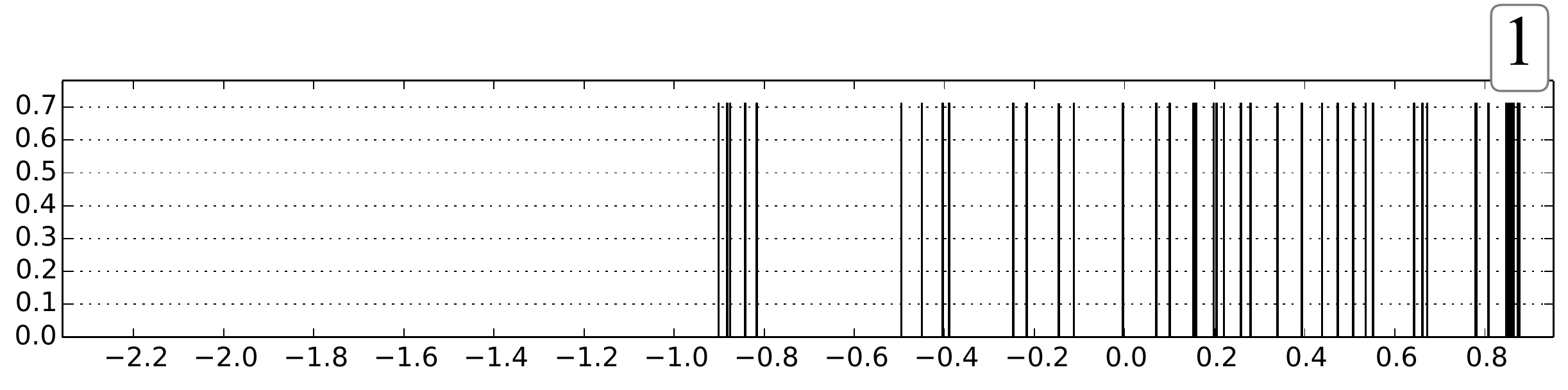} & \piePlotTableI{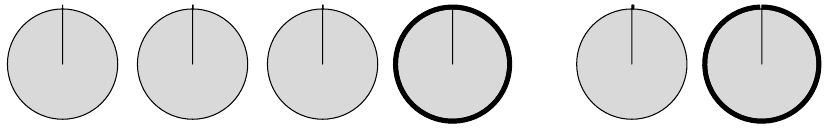} \\
\hline

6 & \textbf{Explore} with \textbf{either} memory \hspace{-0.4cm} &
\pcaPlot{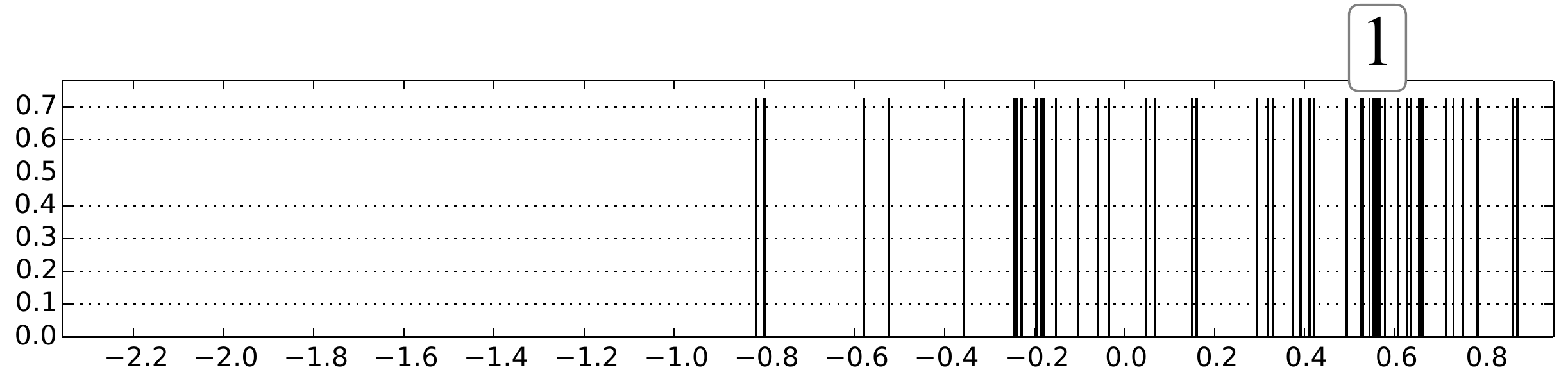} & \piePlotTableI{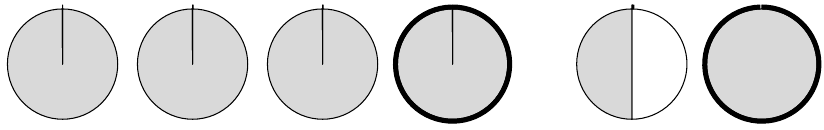} \\
\hline
\hline

7 &
\vcenteredhbox{
	$\hspace{-0.27cm}
	\begin{array}{l}
	\text{Environments 1-6 above and} \\
	\text{one with RTTS opponents} \\
	\text{(Avg. over all environments)}
	\end{array}
	\hspace{-0.4cm}$
} &
\pcaPlot{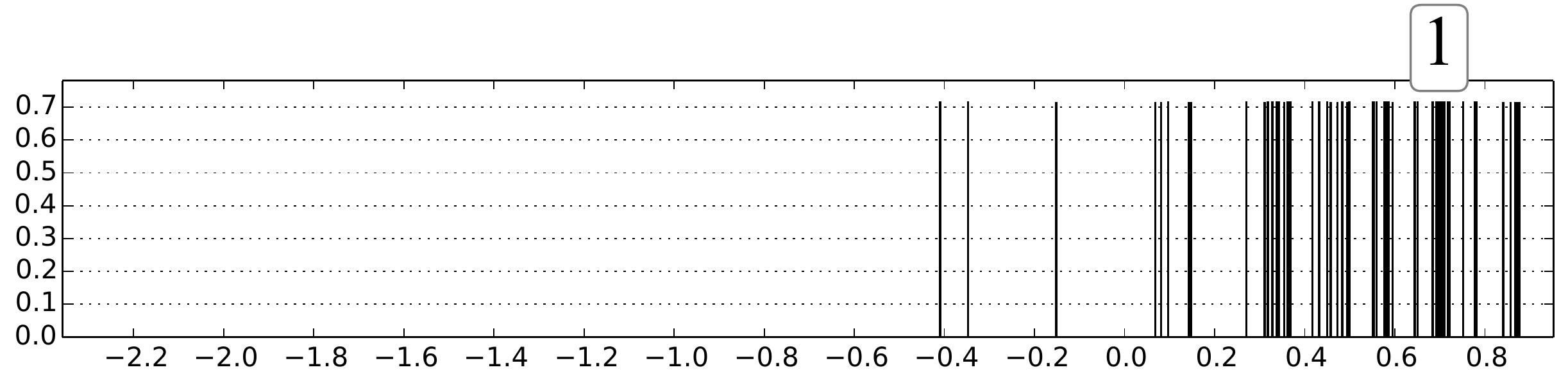} &
\piePlotTableI{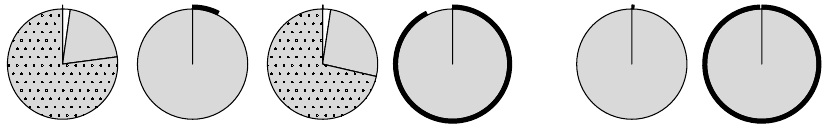} \\
\hline

8 &
\vcenteredhbox{
	$\hspace{-0.27cm}
	\begin{array}{l}
	\text{One opponent from env.} \\
	\text{1-6 and one RTTS opponent} \\
	\text{(Heterogeneous environment)}
	\end{array}
	\hspace{-0.4cm}$
} &
\pcaPlot{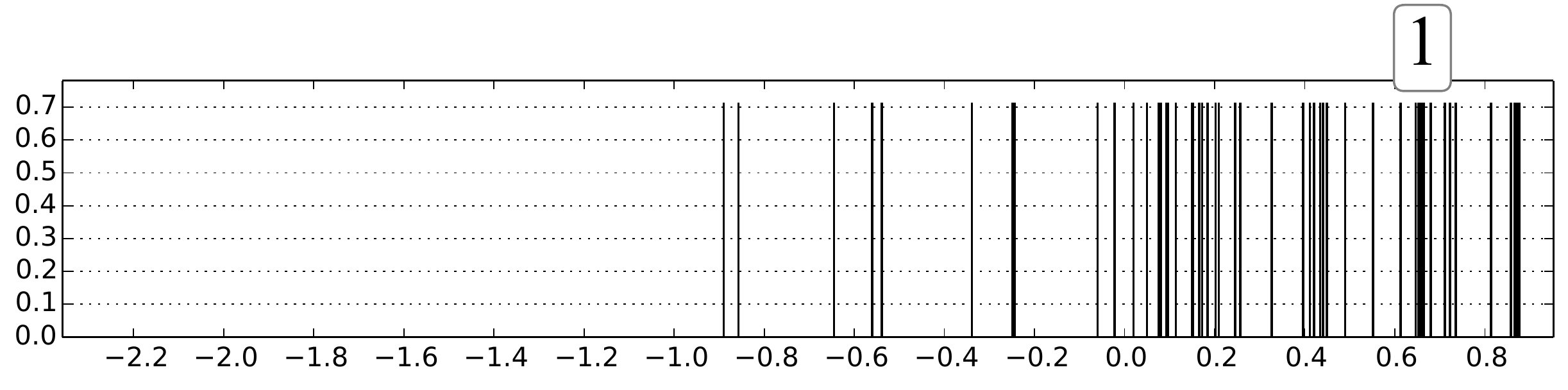} &
\piePlotTableI{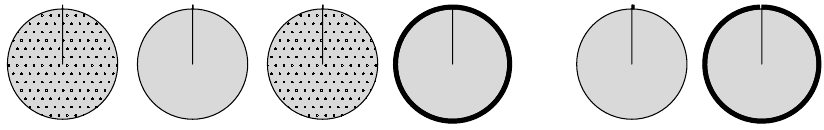} \\
\hline

\pieLegendRow \\

\end{tabular}
}
\caption{
Strategies evolved in sparse environments ($N=20$).
The third column shows all 64 strategies evolved in the environment with the opponents specified in the second column. The $x$-axis is the first principal component (from PCA across all strategies in all sparse environments) and the $y$-axis is the fitness of the strategy. The $x$-axis represents primarily public vs. private memory preference, with public memory use increasing toward the left and private memory use toward the right.
Using the pie-chart format of Figure~\ref{fig:strategy-pie-format}, the fourth column displays sample strategies for each environment, numbered to indicate their positions on the PCA plot.
The black band around each circle indicates the average percentage of time the evolved agent spent in the corresponding state.
Most strategies cluster to the right, indicating that private memory search works well, except in Environments~3 and 4, where the distribution is bimodal because the opponents do not use public memory. In some cases, parts of the strategy do not matter and there is considerable diversity (e.g. in Environment~3).
The best strategies are slightly less than extreme, which allows them to perform better than fixed strategies.
%In many environments, strategies evolved that were more flexible and more complex than the hand-coded ones, making it possible for them to utilize the available information better, and resulting in better performance overall.
}
\label{tbl:evolved-strategies-sparse}
\end{table*}

\begin{table*}[tp]
\centering
\resizebox{\textwidth}{!}{
\begin{tabular}{|c|l|l|l|l|}
\hline
\multicolumn{2}{|l|}{\textbf{Dense Environment}} &
\multirow{2}{*}{\textbf{PCA for All Evolved Strategies}} &
\multirow{2}{*}{\textbf{Sample Strategies}} \\
\cline{1-2}
\textbf{Id} & \textbf{Opponents} & & \\
\hline\hline

\pieOrderDescRow \\  \hline

1 &
\textbf{Exploit} with \textbf{public} memory \hspace{-0.4cm} &
\pcaPlot{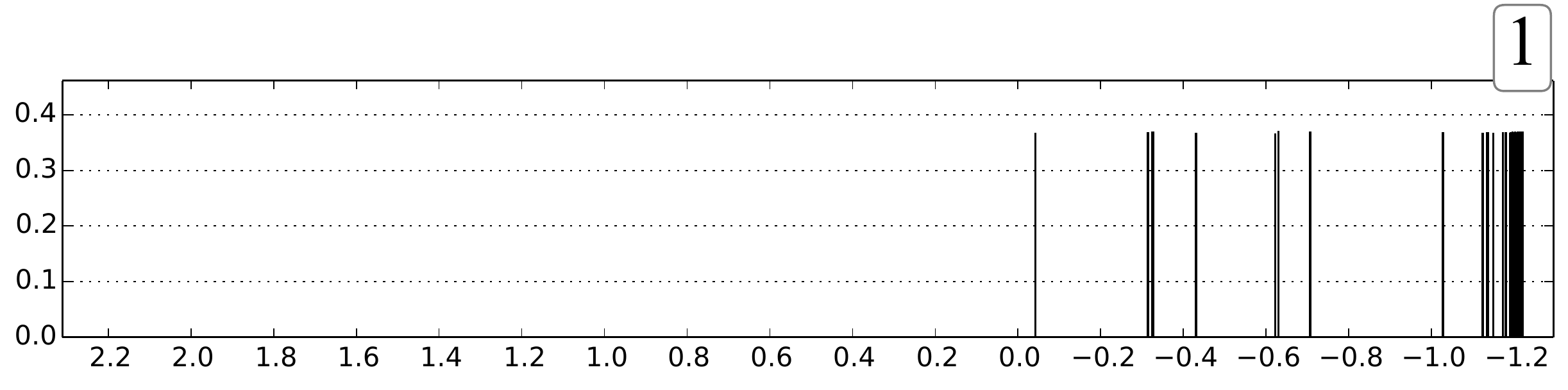} & \piePlotTableI{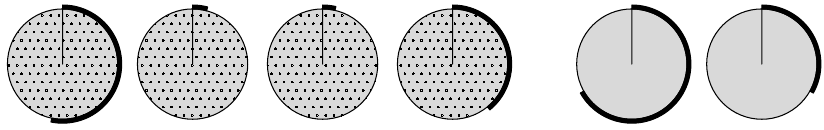} \\
\hline

2 &
\textbf{Explore} with \textbf{public} memory \hspace{-0.4cm} &
\pcaPlot{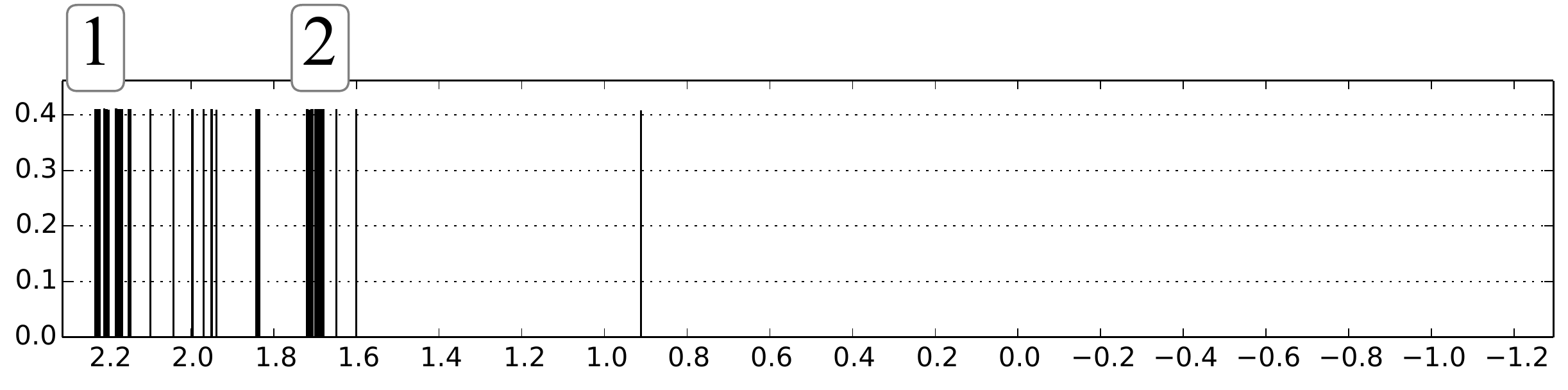} &
\piePlotTableII{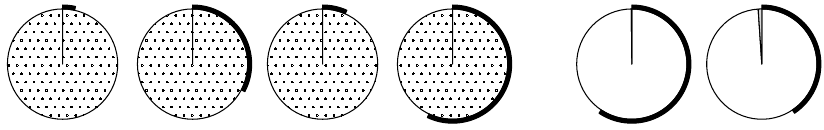}{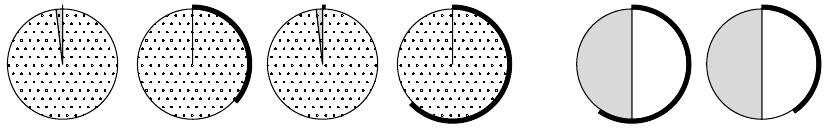} \\
\hline

3 &
\textbf{Exploit} with \textbf{private} memory \hspace{-0.4cm} &
\pcaPlot{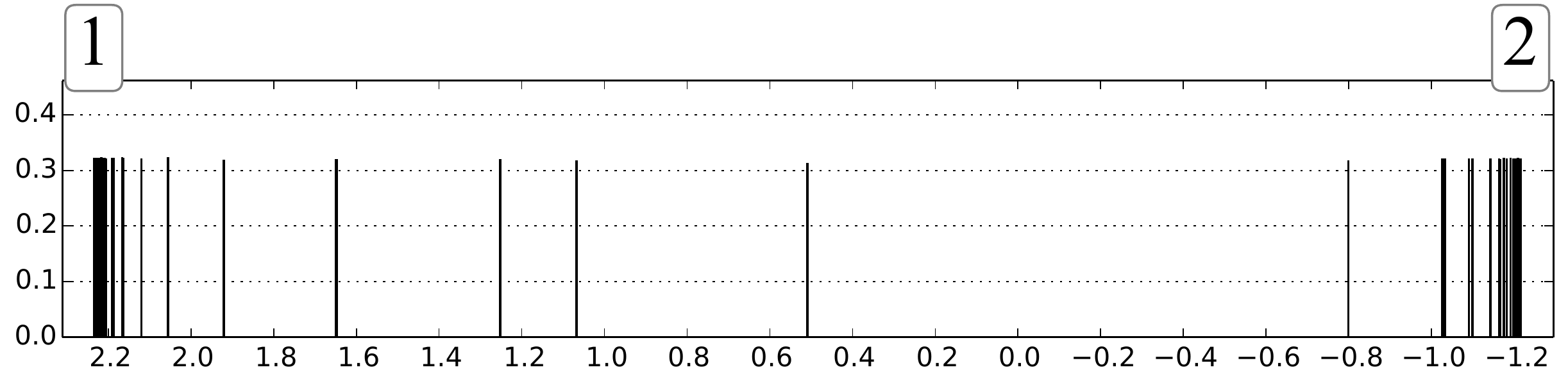} &
\piePlotTableII{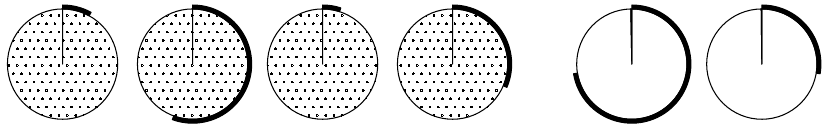}{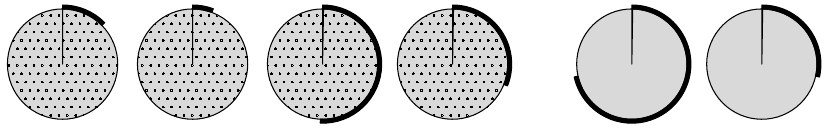} \\
\hline

4 &
\textbf{Explore} with \textbf{private} memory \hspace{-0.4cm} &
\pcaPlot{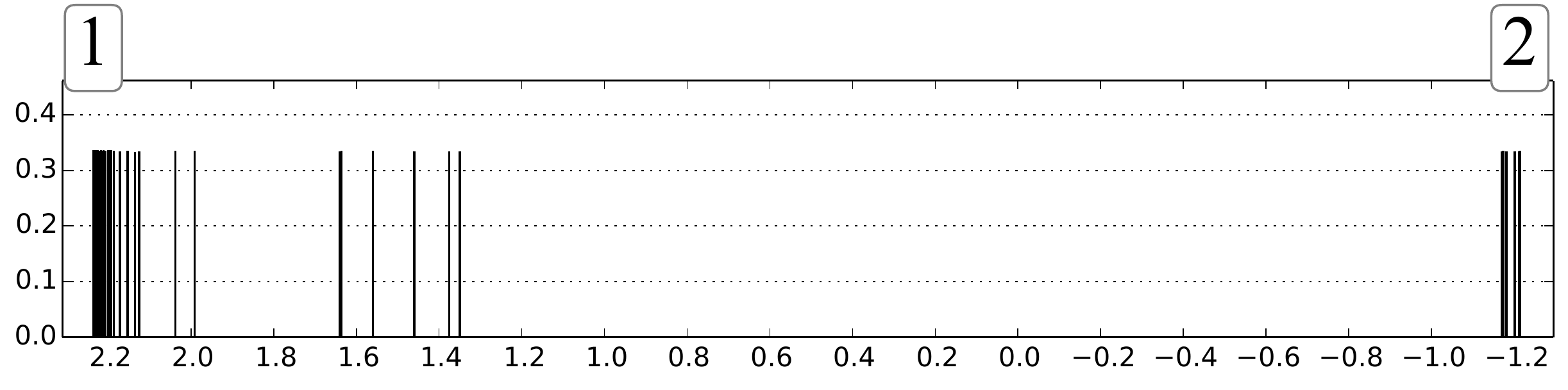} &
\piePlotTableII{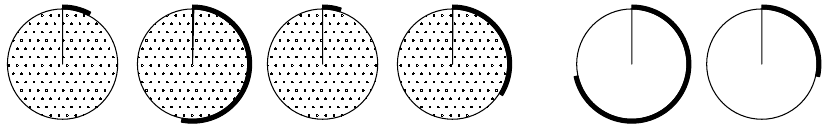}{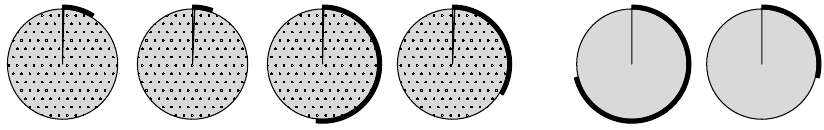} \\
\hline

5 &
\textbf{Exploit} with \textbf{either} memory \hspace{-0.4cm} &
\pcaPlot{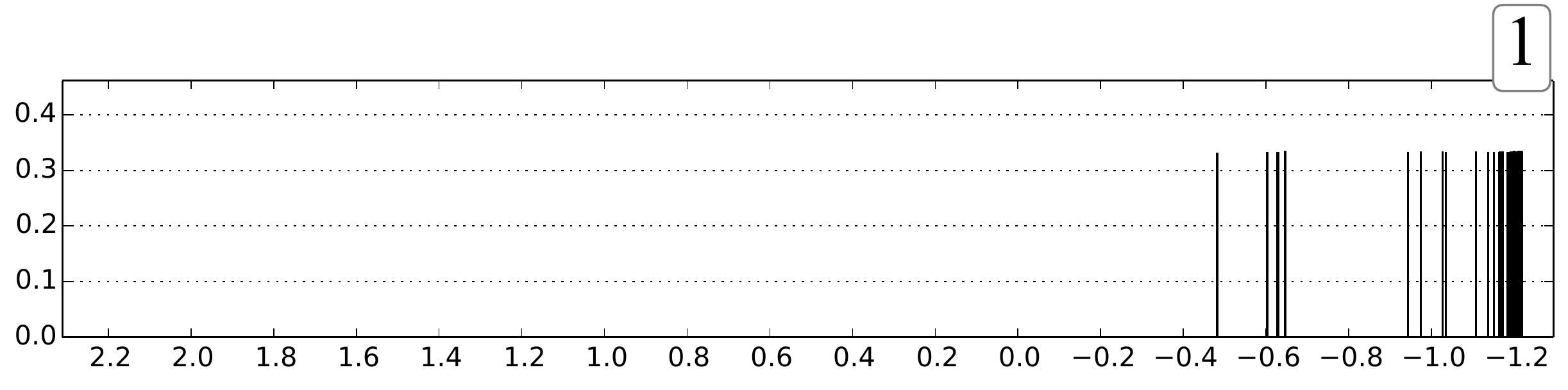} & \piePlotTableI{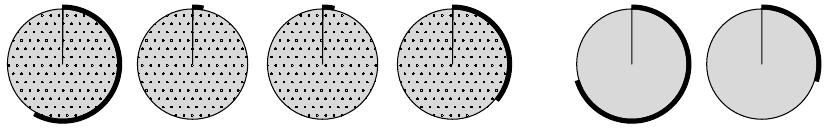} \\
\hline

6 &
\textbf{Explore} with \textbf{either} memory \hspace{-0.4cm} &
\pcaPlot{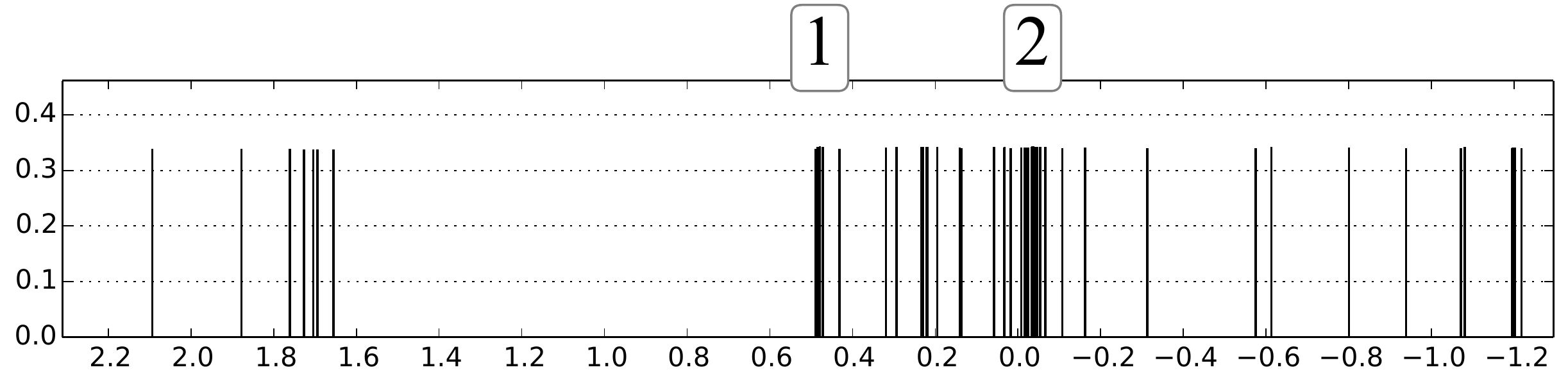} &
\piePlotTableII{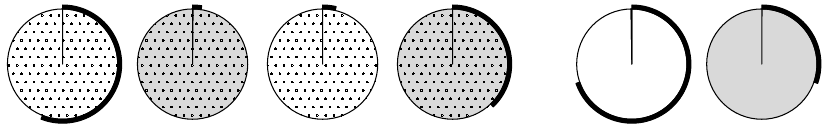}{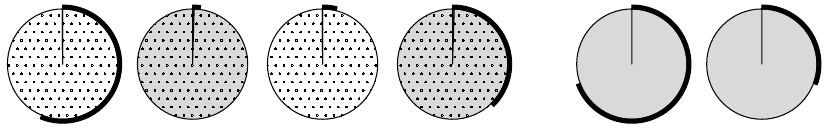} \\
\hline  \hline

7 &
\vcenteredhbox{
	$\hspace{-0.27cm}
	\begin{array}{l}
	\text{Environments 1-6 above and} \\
	\text{one with RTTS opponents} \\
	\text{(Avg. over all environments)}
	\end{array}
	\hspace{-0.4cm}$
} &
\pcaPlot{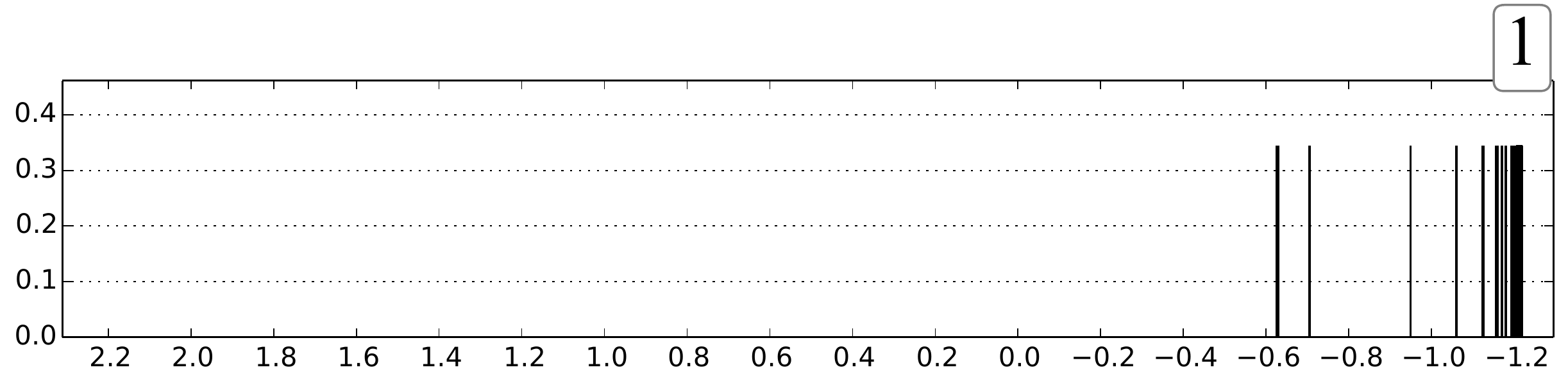} & \piePlotTableI{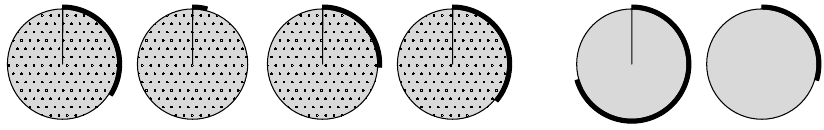} \\
\hline

8 &
\vcenteredhbox{
	$\hspace{-0.27cm}
	\begin{array}{l}
	\text{One opponent from env. 1-6} \\
	\text{and one RTTS opponent} \\
	\text{(Heterogeneous environment)}
	\end{array}
	\hspace{-0.4cm}$
} &
\pcaPlot{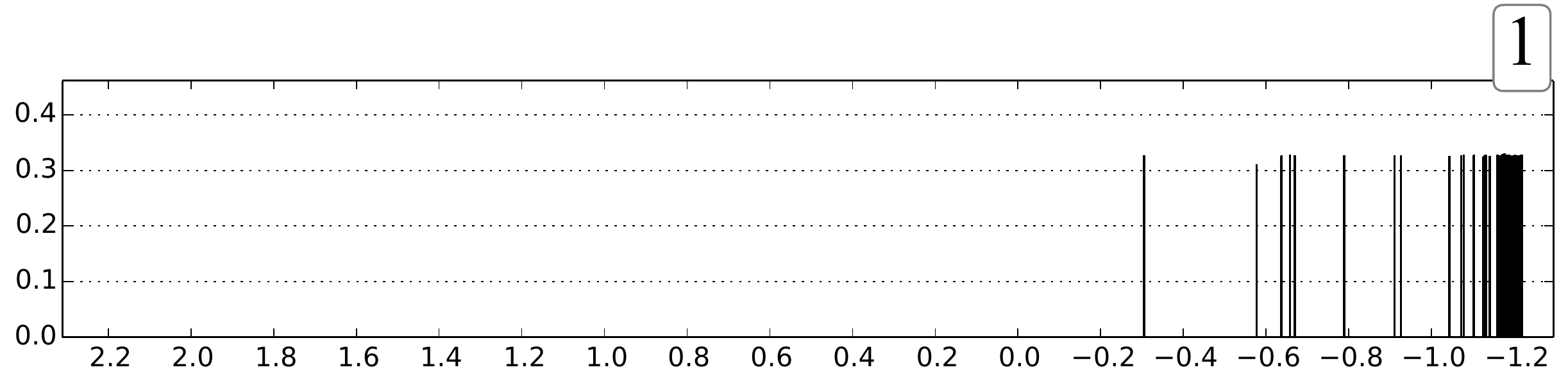} & \piePlotTableI{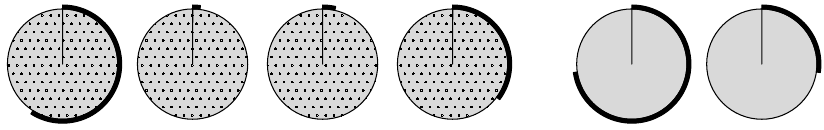} \\
\hline

\pieLegendRow \\

\end{tabular}
}
\caption{Strategies evolved in dense environments ($N=10$). The $x$-axis scale in the PCA plots is inverted so that the shift from public to private memory use is in left-to-right direction as in Table~\ref{tbl:evolved-strategies-sparse}; the $y$-axes have a different scale from Table~\ref{tbl:evolved-strategies-sparse}. In the dense environments, there is less variability due to stronger evolutionary pressure. Memory preference in different environments is similar to that in Table~\ref{tbl:evolved-strategies-sparse}, with the exception of Environment 2, where an opposite memory preference emerged. This effect is investigated further in Figure~\ref{fig:num-prior-visits}.
}
\label{tbl:evolved-strategies-dense}
\end{table*}

%> Large landscapes
Tables~\ref{tbl:evolved-strategies-sparse} and \ref{tbl:evolved-strategies-dense} show distributions of strategies evolved on sparse and dense environments, respectively, with example evolved strategies depicted in pie-charts as in Figure~\ref{fig:strategy-pie-format}.
To visualize the distribution of the 64 evolved strategies, they were first represented as 20-dimensional vectors and their dimensionality was then reduced to one using PCA (separately for sparse and dense environments). In the dense environments, the first principal component captured 83\% of the variance, and in the sparse environments, 39\%, whereas the second principal component captured 8\% and 16\%, respectively, suggesting that one-dimensional visualization is indeed meaningful.

The first PCA dimension is the $x$-axis of the plots in the third column of Tables~\ref{tbl:evolved-strategies-sparse} and \ref{tbl:evolved-strategies-dense}. Note that the $x$-axis scale is the same for all sparse environments, but different from the axis for the dense environments. The $y$-axis of the PCA plots indicates the fitness of the strategies.
Interestingly, in both sparse and dense environments the $x$-axis corresponds to a preference for using public or private memory: The leftmost strategies across all plots have the highest public memory use, whereas the rightmost ones use private memory the most. This tendency can be clearly seen e.g. in the two sample strategies for Environment~4 in Table~\ref{tbl:evolved-strategies-sparse}.

Several interesting observations can be made based on Tables~\ref{tbl:evolved-strategies-sparse} and \ref{tbl:evolved-strategies-dense}.
First, most strategies cluster on one side of the PCA plot, except for Environments 3 and 4, where a bimodal distribution is observed. These two environments are the only ones where the opponents uses only private memory. Since the opponents never use public points, public memory effectively becomes just another private memory for the evolved agent, and a bimodal distribution results. This result is very similar to the observation in Section~\ref{sec:simEnv} where public and private memory use resulted in similar performance when the opponents used only private memory.

Second, the distribution of strategies evolved in most of the remaining environments is biased toward strategies that prefer private memory over public memory. Thus, evolution discovered that it is good to be different from competitors. This result also parallels the performance comparison of manual strategies in Figure~\ref{fig:evalComparisonManual}.

Third, in Environment 2 in Table~\ref{tbl:evolved-strategies-dense}, where opponents explore with public memory, a surprising opposite effect is seen: public memory is preferred over private memory. Not only is this result counterintuitive, it is also the opposite of that in its sparse counterpart.

\begin{figure}[t]
\centerline{\includegraphics[width=\columnwidth]{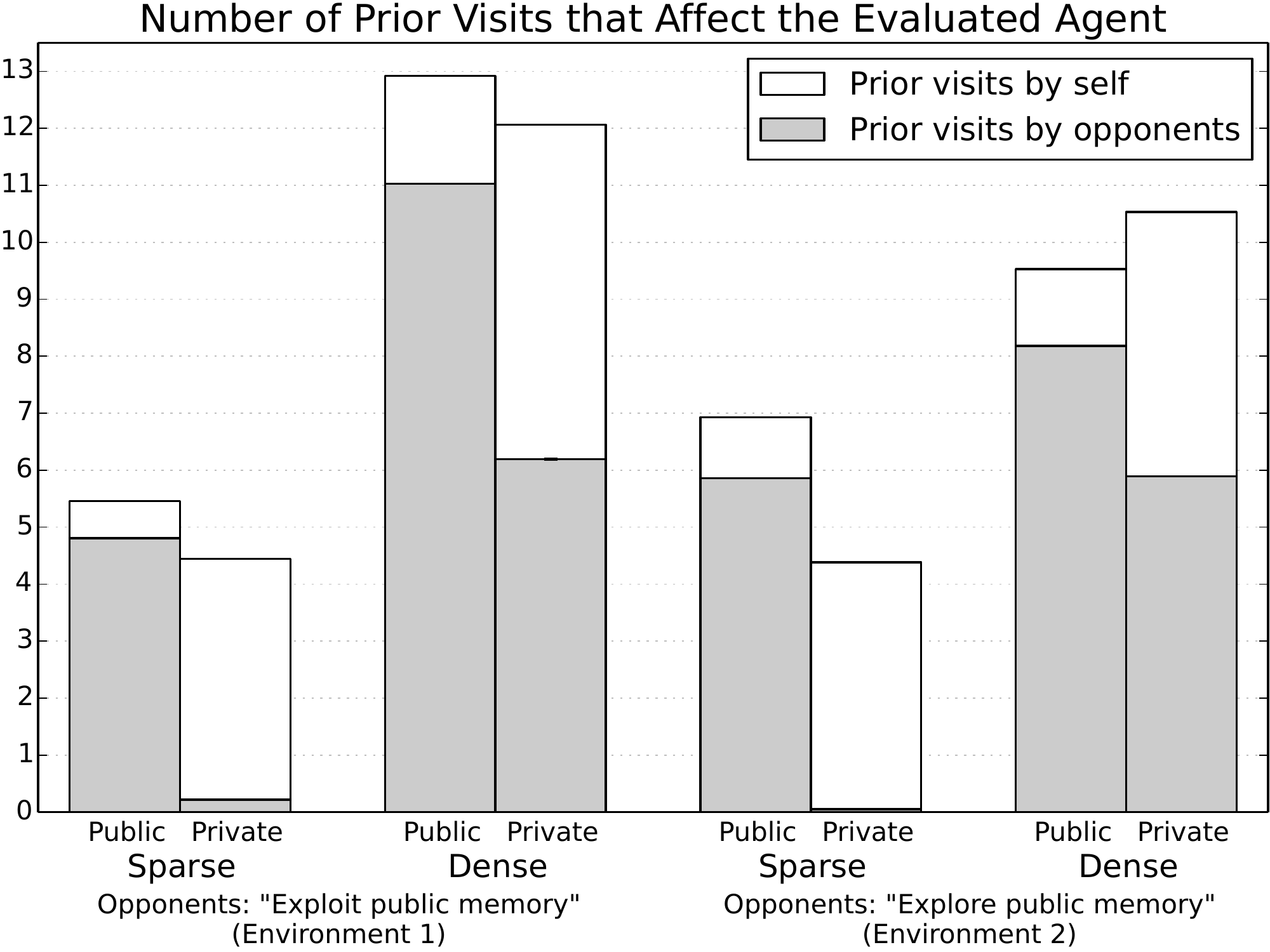}}
\caption{The number of prior agent visits within the flocking radius of the evaluated agent in Environments~1 and 2 in Tables~\ref{tbl:evolved-strategies-sparse} and \ref{tbl:evolved-strategies-dense}. Agent visits cause the landscape to sink due to crowding, and lead to lower fitness.
The bars labeled Public represent the average number of prior visits when the evaluated agent uses one of the 64 evolved strategies that prefer public memory the most (i.e. the leftmost 64 strategies in the PCA plots of all environments in Tables~\ref{tbl:evolved-strategies-sparse}, for the bars labeled Sparse, and \ref{tbl:evolved-strategies-dense}, for the bars labeled Dense. Similarly, the bars labeled Private show the average number of visits with the rightmost 64 strategies in the same tables.
The bars' colors, white and gray, indicate prior visits by the evaluated agent (i.e. ``self'') or by its opponents, respectively.
Switching from private strategies to public ones increases the total number of prior visits in three out of the four environments, but not in dense environment~2. As a result, evolution selects for a public memory strategy in this case demonstrating that it can find effective strategies that would be difficult to discover by hand.
}
\label{fig:num-prior-visits}
\end{figure}

This effect can be explained by measuring the number of prior visits to a given area by the agent and its opponents. As Figure~\ref{fig:num-prior-visits} shows, the number of prior visits that affect the evaluated agent is usually higher when the agent uses a public-memory strategy than when it uses a private-memory strategy, and therefore evolution usually favors private memory. However, the opposite is true for dense environment 2: The public-memory strategy actually results in fewer prior visits. The reason becomes clear when the source of prior visits is considered. When the evaluated agent uses private memory, it is affected by many of its own prior visits, but manages to avoid opponent's prior visits relatively well. When it switches to using public memory, it reduces its own prior visits, but also increases the opponent's visits. Such a trade is usually detrimental, but not so in the dense environment 2. Because the environment is dense and the opponents are exploring, even when the agent is using private memory, the opponents make many prior visits already by chance. Switching to public strategy therefore does not increase the opponent visits much, but it does reduce self-visits significantly. The net effect is therefore beneficial, and evolution will select for public-memory strategy for dense environment 2.

Fourth, the agents encounter different states in dense and sparse environments. This result can be seen by observing the black bands around the circles of sample strategies in Tables~\ref{tbl:evolved-strategies-sparse} and \ref{tbl:evolved-strategies-dense}, which represent the average percentage of time spent in each $S_{1}$ and $S_{2}$ state. Across all evolved strategies in dense environments, the $S_{1}$ state where public and private memory have low fitness (i.e.\ the first circle) is encountered 34\% of the time on average, whereas in sparse environments only 0.011\%. Similarly, in dense environments the evolved agents spent 69\% of their time in the $S_{2}$ state where the new point has low fitness, but only 0.593\% in sparse environments. The reason is that the fitness landscape sinks more in dense environments, and by the end of the 100-step simulation, all points in the search space have low fitness. In contrast, in sparse environments, there are often points with high fitness that the agents can visit, and the agents evolve to do so. Therefore, evolved agents spend most of their time in high-fitness states in sparse environments and low-fitness states in dense environments.

%Fourth, the black bands around the circles of sample strategies, which indicate the average percentage of time spent in the $S_{1}$ and $S_{2}$ states that correspond to each circle, hint that there is a clear difference in the states encountered by agents in dense and sparse environments, which is confirmed by the average percentages across all evolved strategies: In dense environments, the $S_{1}$ state where public and private memory have low fitness (i.e. the first circle) is encountered 34\% of the time on average whereas in sparse environments only 0.011\%. Similarly, in dense environments the evolved agents spent 69\% of their time in the $S_{2}$ state where the new point has low fitness, but only 0.593\% in sparse environments. The difference in observing low and high fitness states between sparse and dense environments is not surprising since high agent density leads to sinking of the whole fitness landscape, with all points in the search space having low fitness by the end of the 100-step simulation, whereas in sparse ones, there are often points with high fitness that the agents can visit, which results in agent strategies that evolve to do so. Therefore, those evolved agents spend most of their time in high-fitness states in sparse environments, unlike in dense ones.

Fifth, a comparison of the PCA plots across the two tables indicates that the strategies evolved in sparse environments vary more than those evolved in dense ones. For example, in Environment 3 in Table~\ref{tbl:evolved-strategies-sparse} the evolved strategies are dispersed much more than their counterparts in Table~\ref{tbl:evolved-strategies-dense}. It is interesting to analyze why.  First, note that strategies located on the left side (i.e.\ those that use public memory more often than private memory) mostly differ in which actions they take when fitness of public memory is low (i.e.\ the first two circles of sample strategies 1, 2, and 3), and when the points they visit have low fitness (i.e.\ the fifth circle).  Note further that such states are visited only rarely (i.e.\ less than 1\% of the time), as indicated by very little black band around the circles. The reason is that only the agent itself changes public memory (not its opponents, which exploit with private memory), and it places only high-fitness points in it. Therefore, public memory always has a high fitness, and as a result, the first two states are rarely encountered and the corresponding parts of the strategy (represented by the first two circles) are rarely used. Also, since low-fitness points are usually not visited in sparse environments, the fifth circle is rarely used.  Similar observations can be made for the strategies on the right side (i.e.\ those that use private memory more often, such as sample strategies 4 and 5): They vary mostly in states where private memory has low fitness (i.e.\ the first and third circle) as well as when visited points have a low fitness (i.e.\ the fifth circle), and these are indeed states they rarely visit. Thus, changes to those parts of the strategy do not affect the evolved agent's performance, and evolution results in diverse solutions for them.  In contrast, in dense environments all states are encountered at least 9\% of the time on average: All parts of the strategy are therefore useful, and there is less variance.

%Fifth, a comparison of the PCA plots across the two tables indicates that in general the strategies evolved in sparse environments vary more than those evolved in dense ones.
%% A likely explanation is that there is more evolutionary pressure in dense environments, which leads to lower genetic diversity >>> citation.
%For example, in Environment~3 in Table~\ref{tbl:evolved-strategies-sparse} the evolved strategies at the left side, which use public memory more often than private memory, are dispersed much more than their counterparts in the corresponding environment in Table~\ref{tbl:evolved-strategies-dense}. The pie charts for some of those strategies in sparse environment~3 (e.g. the sample strategies numbered 1, 2, and 3) indicate that the main differences between them are in what actions they take when fitness of public memory is low (i.e. the first two circles on each row), and when the point they visit has low fitness (i.e. the fifth circle).
%%Effectively, they all represent the strategy ``exploit public memory, place visited points in public memory''.
%When those three sample strategies are used in the environment where they were evolved, the states that correspond to those three circles are exactly the ones that are rarely encountered (i.e. less than 1\% of the time), which is indicated by the lack of a black band around those circles, unlike the third, fourth, and sixth circles.

In Environment~3, opponents exploit private memory, and place the new points they find in private memory, which makes the evolved agent the only one (among the eight agents in the simulation) that changes public memory. Since each one of sample strategies 1, 2, and 3 places new high-fitness points in public memory (i.e. the sixth circle), public memory always has high fitness. Therefore, the first two states (which together represent public memory having low fitness) are not observed with those strategies. Thus, the corresponding parts of the strategy (i.e. the first two circles and rows in the $S_{1}$ table) are not used during simulation.
In contrast, if the evolved agent uses private memory instead of public, as in sample strategy numbered 4 and 5,
the unused states are different: The first and third circles, which together represent the state of private memory having low fitness, do not get used.
%, which places new high fitness points in private memory instead,
%with sample strategies 1, 2, and 3.
Since low fitness points are usually not visited in sparse environments, the parts of the strategy for new low-fitness points (i.e. the fifth circle and the first row of the $S_{2}$ table) are not used either. Thus, changes to those parts of strategy do not affect the evolved agent's performance, and during evolution strategy variants with differences in those unused parts do arise, which leads to higher diversity in the evolved strategies in sparse environments. On the other hand, in dense environments all states are encountered at least 9\% of the time on average, which makes most states useful in each environment for the evolved strategies. Since most parts of the strategies are used in dense environments, there is relatively lower variance in evolved strategies compared to sparse environments.
%causes the actions on low-fitness public points to be unimportant to the agent.
%Fifth, in some environments, there is a diverse set of strategies that all perform well.
%In such cases, there are components of the strategy that do not make a difference.
%For example, in Environment~3 in Table~\ref{tbl:evolved-strategies-sparse} the opponents exploit private memory; the evolved agents at the left side (numbered 1, 2, and 3) use high-fitness points in public, and the actions on low-fitness public points do not matter.
%... gives a hint for explaining the higher variance in sparse environments compared to dense ones

% The reason for N=20 causing a diversity level higher than with N=10, is that in the latter case, pretty much all states are encountered, hence matter, leading to mutations anywhere in the genome and in turn changes anywhere in the strategy to affect performance, whereas in the N=20 case, parts of the genome affecting the probabilities of actions in mostly unused states, such as (low pub., low priv.) for S1, and (point: low fitness) for S2, are free to mutate and drift, leading to higher diversity in the evolved strategies.

% In fact, when we limit the PCA to only those probabilities in the states that are consistently encountered by the agent (i.e. the last row of S1 and S2 tables: (high pub., high priv.) for S1 and (point: high fitness) for S2), then the first dimension of PCA explains 99.998\% of the variance (instead of the 39\% we now have for N=20), and we end up with PCA plots that look like the attached one, with very little variance.

\begin{figure*}
\centering
\subfloat[Sparse environments]{
\includegraphics[width=0.9\textwidth]{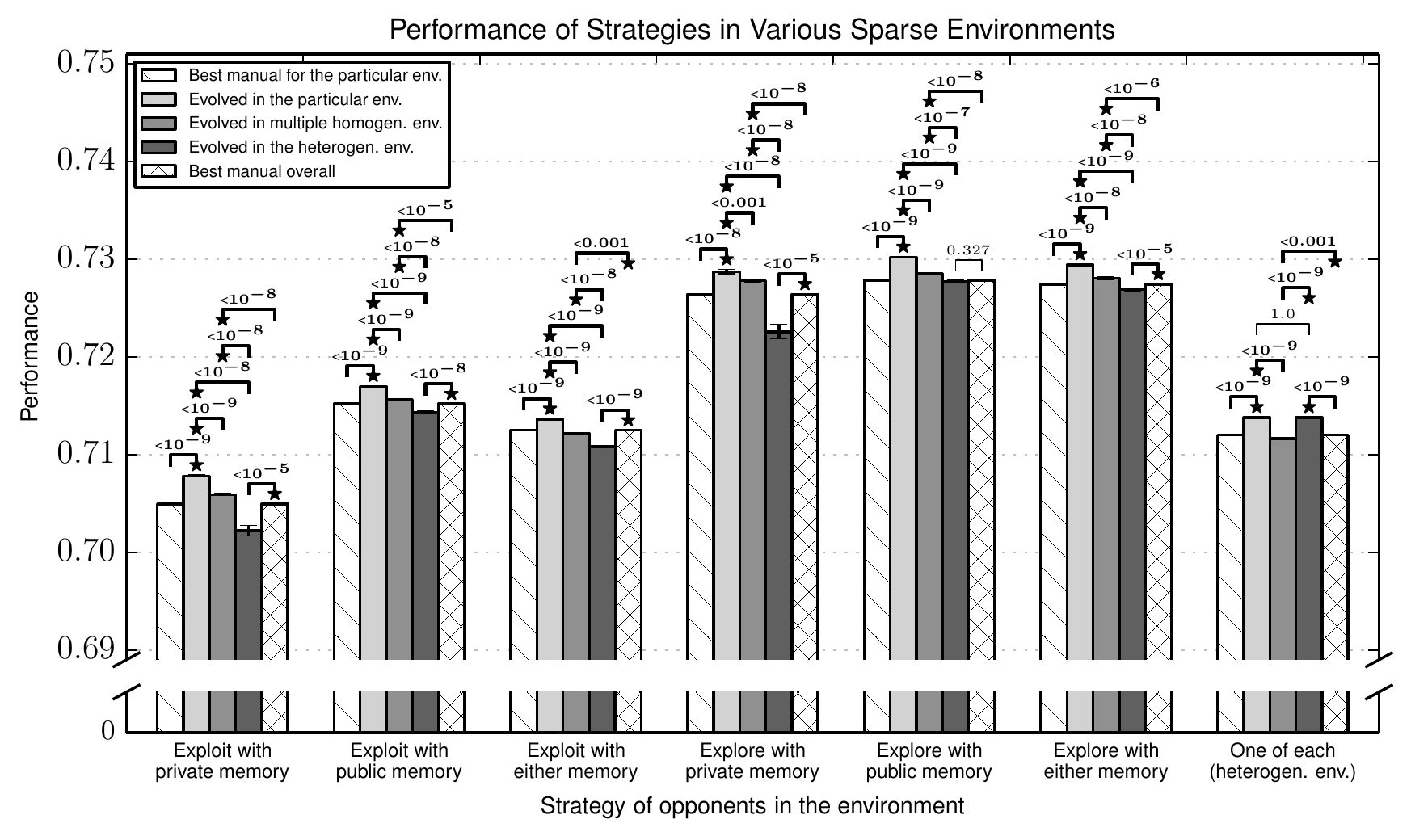}\label{fig:evalComparisonEvoSparse}
}\\
\subfloat[Dense environments]{
\includegraphics[width=0.9\textwidth]{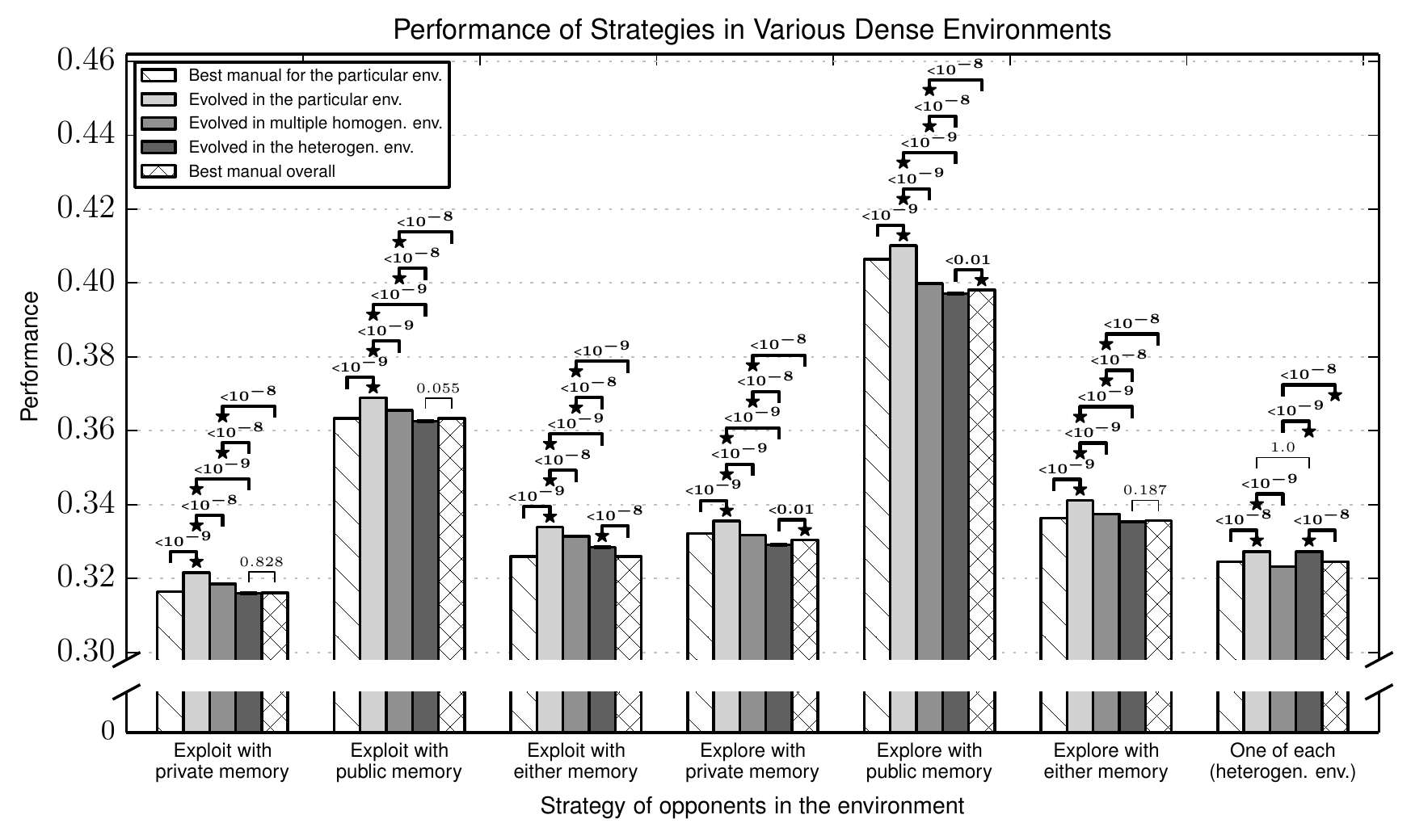}\label{fig:evalComparisonEvoDense}
}
\caption[]{Performance comparison among evolved strategies as well as the best manual strategy from Figure~\ref{fig:evalComparisonManual} in sparse ($N=20$) and dense ($N=10$) environments.
Statistical significance is estimated between averages over 64 evolution runs.
Significant differences are indicated by stars.
The same six environments are included as in Figure~\ref{fig:evalComparisonManual}, as well as a seventh one where each opponent had a different hand-coded strategy, where the strategies whose performance is shown with the second and fourth bars in the plot are identical. Hence the 1.0 $p$-value between those two bars. Strategies evolved for each environment separately perform the best in all cases, while the general strategies evolved in multiple environments are better than the best manual overall strategy in almost all cases. Machine discovery is therefore a powerful approach to develop CMAS strategies.
}
\label{fig:evalComparisonEvo}
\end{figure*}

Sixth, the best strategies are not perfectly extreme, unlike the fixed hand-coded ones, but often contain small slivers of probability for alternative actions. Such small differences allow them to perform better.
As can be seen in Figure~\ref{fig:evalComparisonEvo}, in each environment the evolutionary optimization resulted in a strategy that performs at least as well as the best manual strategy.

Thus, the results verify the hypothesis that custom-designed strategies are usually more successful than generic ones. An interesting question is: Is there a general strategy that works well on all environments?

\subsection{Evolving General Strategies}
\label{sec:heterogeneousEnv}

%> Multiple homogeneous environments
Two ways of evolving general strategies were tested: evolving in multiple homogeneous environments and in a single heterogeneous environment.
In the first approach, seven homogeneous environments were used, consisting of the six environments above, and an environment with seven opponents that use an adapted RTTS strategy (which will be described in the next section). During evolution the fitness of each evaluated strategy was calculated by averaging the fitness score across those environments.
The main disadvantage of this approach is that it takes a very long time: each strategy must be evaluated in seven environments in the using a total of 700 simulation steps.

%> Heterogeneous environment
The approach that uses a single heterogeneous environment avoids this problem.
There is one opponent of each type in this environment, allowing the agent to interact with various types of opponents at once. Therefore, evaluation only requires one environment and 100 simulation steps. As in previous experiments, there are seven opponents but now each of them comes from a different homogeneous environment.

The strategies evolved using these two approaches can be seen in Tables~\ref{tbl:evolved-strategies-sparse} and \ref{tbl:evolved-strategies-dense} as Environments 7 and 8. Private memory was mostly preferred over public memory in both sparse and dense environments. With the homogeneous approach, strategies evolved in the sparse environment exploited private memory when private memory had high fitness (Environment 7 in Table~\ref{tbl:evolved-strategies-sparse}); otherwise, their behavior varied, including exploring private memory while still sometimes exploiting it, as shown in the sample strategy. When the environments were dense (Environment 7 in Table \ref{tbl:evolved-strategies-dense}), this approach evolved strategies that mostly explored, but also rarely exploited with private memory.

Similarly, in the heterogeneous approach (Environment 8), the evolved strategies in the sparse environment exploited with private memory when private memory had high fitness, but also explored when private memory had low fitness. On the other hand, the strategies evolved in the dense heterogeneous environment always explored with private memory. The likely reason is the same as in Section~\ref{sec:expEvolveParticular}: there are opponents that use public memory, making it less beneficial to use.

Performance of the best strategies evolved using the two general approaches was compared to that of the best strategy evolved specifically for that particular environment, as well as to the manual strategy that performed best in that environment. The results can be seen in Figure~\ref{fig:evalComparisonEvo}. In both sparse and dense environments, the performance of the strategy that was evolved for that particular environment was always the best of the evolved strategies (with $p$-value $< 10^{-9}$ compared to the strategy with the second highest mean in most environments, and $< 10^{-3}$ in the rest), and that was always followed by the single strategy that was evolved in multiple homogeneous environments, which in turn always performed better than the one evolved in the single heterogeneous environment (with $p$-value $< 10^{-7}$ in all environments). Interestingly, the performance of the single heterogeneous environment was on average within 1\% of that of multiple homogeneous environments, even though it required one-seventh of the evolution time.
Thus, evolution in a heterogeneous environment is an elegant and effective approach to finding general strategies.

Overall, the similarity in performance between the different learning approaches suggests that it may be possible to evolve a single strategy that is effective in various environments, although the very best results are obtained by customizing the strategy to each particular environment separately.

\subsection{Real-Time Tree-Search Agents}
\label{sec:rttsAgents}

In order to highlight how different CMAS problems are from conventional search problems, a real-time tree search (RTTS) algorithm was devised for the $N\!K$ fitness landscape.
RTTS is real-time in the sense that it does not perform the whole search offline like A* does, but instead alternates between planning and execution phases by performing a limited look-ahead search at each state before selecting an action and moving to a new state. In this respect, this algorithm is similar to e.g. Real-Time A* (RTA*) \cite{Korf:ai90}, a well-known search method in the single-agent tradition. Further, at each step RTTS, like RTA*, performs a full exploitation search from the current point, i.e. considers all successor states reachable by exploitative search actions. In contrast, CMAS also includes exploratory search actions, which can potentially reach any point in the search space.

The reason for defining and employing RTTS instead of simply using RTA* is that the search domain of $N\!K$ fitness landscapes differs in several ways from those for which RTA* was designed: (1) The goal is not to reach a certain point in the search space as in RTA*, but rather to follow a path that yields as much fitness as possible; (2) Since there is no goal point, there cannot be a heuristic to calculate the cost of reaching a goal point; (3) Avoiding loops is not a concern as long as the revisited points have high fitness; (4) The search space is dynamic due to flocking of agents, which makes it less useful to keep a hash table of observed states and their estimated costs for returning to those states. Thus, RTTS can be seen as an adaptation of RTA* to CMAS problems.

The RTTS algorithm works as follows.
Given an agent's state $s$ (in this case, the last point the agent has visited), a score is calculated for each successor state $s'$. 
Since only exploit actions are considered, a successor of a state is equivalent to a neighbor of a point in the space. Each point has one neighbor $s'_i$ per dimension $i$ ($1\leq i\leq N$), obtained by flipping the bit of that dimension in point $s$. For each neighbor $s'_i$, RTTS carries out a look-ahead search starting from that state, and calculates the score of $s'_i$ by summing the fitness of that point and the maximum fitness among those of the successors of $s'_i$. The agent's next move is chosen as the exploitation action that results in the state with the maximum score among all $s'_i$.

\begin{figure}[t]
\centering
\includegraphics[width=\columnwidth]{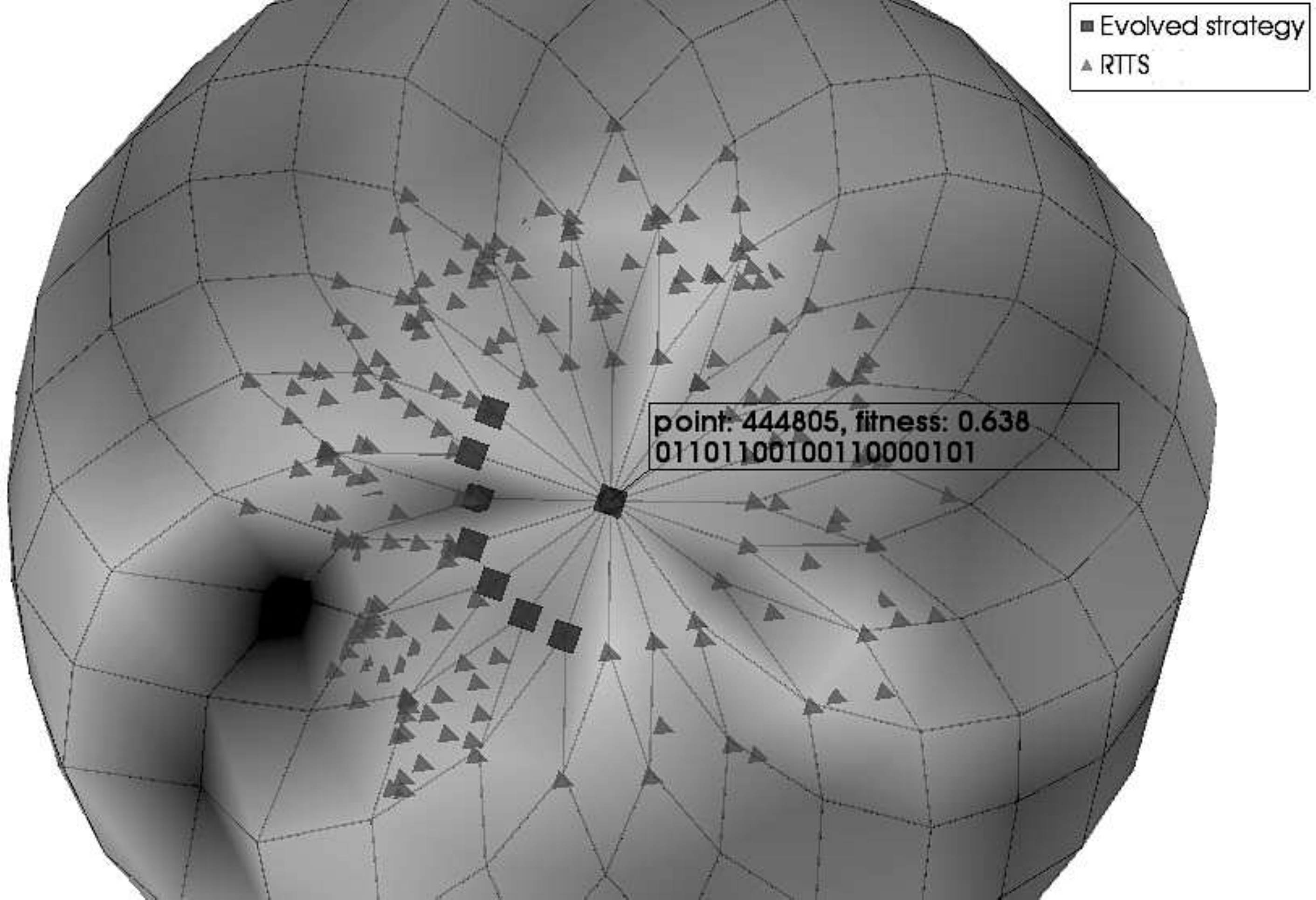}
\caption[]{
Point evaluations with RTTS and an evolved strategy.
The environment is 20-dimensional and shown in a spherical visualization where elevation and brightness represent fitness, and distance from the center point approximates the Hamming distance from it (see Appendix~A for details of this visualization).
The 211 search points that the RTTS agent evaluates in a single time step in order to determine which action to take are shown as triangles. All of them are within two steps of the starting point for the search, shown at the center. In contrast, evolved strategies perform eight point evaluations on average (shown as squares), underscoring how different the CMAS strategies are from classical single-agent search methods.
}
\label{fig:RTAConsideredPoints}
\end{figure}

In fact, what the RTTS agent does for each $s'_i$ is identical to what the agent does for $s$ itself. Thus, the agent's search can be described as fixed-depth tree search. When the depth of this search tree is set to two, the points that RTTS evaluates consist of point $s$ itself, all neighbors $s'_i$ of $s$ ($1\leq i\leq N$), and all neighbors of all $s'_i$. The number of these points is $1 + N(N+1)/2$, which amounts to 56 and 211 points for $N=10$ and $N=20$, respectively. Therefore, to keep the number of evaluations at a reasonable level, the RTTS search depth was limited to two (Figure~\ref{fig:RTAConsideredPoints}).

\begin{figure*}
\centering
\subfloat[Sparse environments]{
\includegraphics[width=0.9\textwidth]{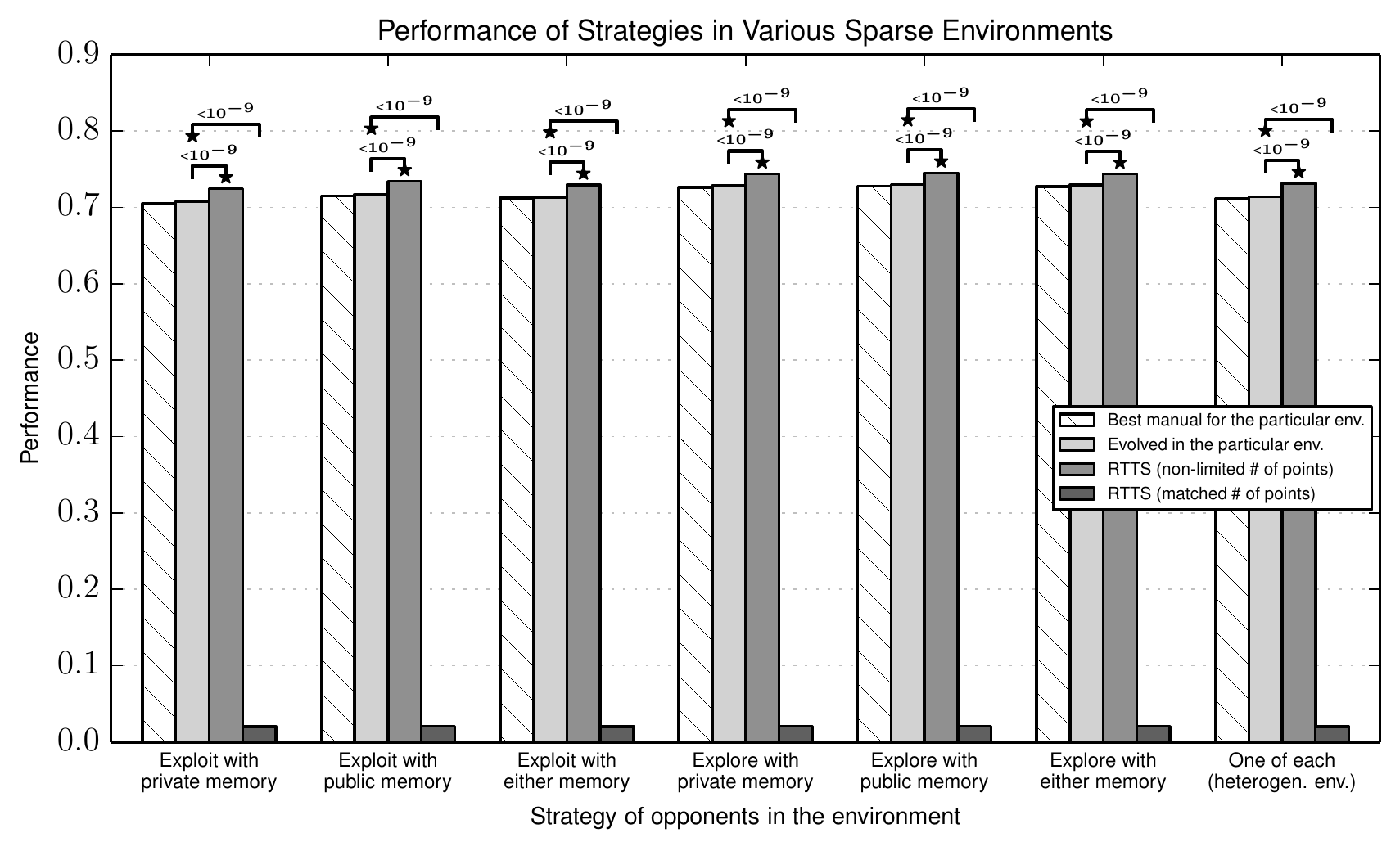}
\label{fig:evalComparisonRTTSSparse}
}\\
\subfloat[Dense environments]{
\includegraphics[width=0.9\textwidth]{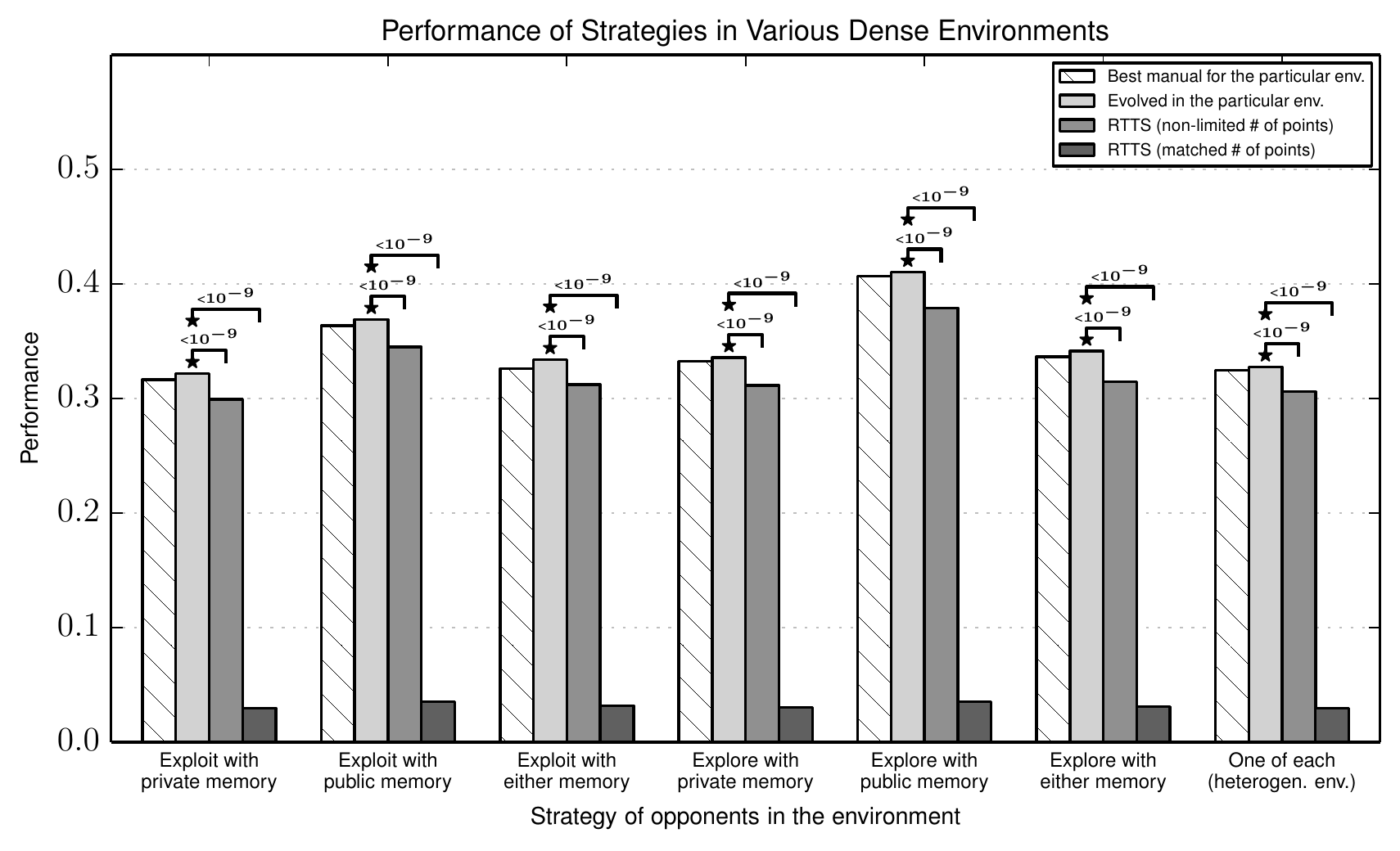}
\label{fig:evalComparisonRTTSDense}
}
\caption[]{Performance of the RTSS strategy compared to the best manual strategy (Section~\ref{sec:simEnv}) and the best evolved strategy (Section~\ref{sec:expEvolveParticular}), in sparse ($N=20$) and dense ($N=10$) environments.
The non-limited version was based on a complete 2-ply lookahead at each step, whereas the matched version evaluated the same number of points as the other methods. The non-limited version is comparable to the other methods, but the matched version is much worse. The assumptions of RTSS do not hold in CMAS problems, which thereby require a different approach.
}
\label{fig:evalComparisonRTSS}
\end{figure*}

The results, compared to best evolved CMAS strategies, are shown in Figure~\ref{fig:evalComparisonRTSS}.
Note that RTTS with search of depth one would only reach the nearest neighbors, amounting to the \emph{always exploit} strategy, with the small difference that all neighbors are considered instead of stopping at the first neighbor that improves over the current point. At depth two, however, RTTS is a distinctly different strategy from those considered so far; it is a traditional single-agent search method adapted to the CMAS setting.

The differences between RTTS and CMAS methods are clear in the results.
The evolved agents as well as the manual CMAS strategies evaluate only eight points per step on average (Figure~\ref{fig:RTAConsideredPoints}). Thus, they are significantly more economical than RTTS in a high-dimensional landscape (i.e. 26 times more in the sparse environments and seven times more in the dense ones).
To make RTTS more comparable with the CMAS methods, it is possible to match the number of points it considers with that of the CMAS methods, by limiting it to one action every 26 time steps in the sparse environments and every seven time steps in the dense ones. As can be seen in Figure~\ref{fig:evalComparisonRTSS}, under such limited resources, RTTS makes very little progress. Whereas the CMAS methods are designed to proceed with the information gained from only a few points, RTTS expects to see the entire depth-2 search tree before making a decision.

Interestingly, even without the resource limitation, RTTS is still not better than the CMAS methods (Figure~\ref{fig:evalComparisonRTSS}). The reason is that it is constantly mislead by the dynamic landscape:
The fitness value of a point that looked promising during the look-ahead may diminish once the agent gets there, and it may miss points whose value increased.

Thus, CMAS problems are different from classical single-agent search problems, and can be solved better by methods designed for such problems in mind, such as those proposed in this article.

\subsection{Visualizing Evolved Strategies}

\begin{figure}[t]
\centering
\includegraphics[width=\columnwidth]{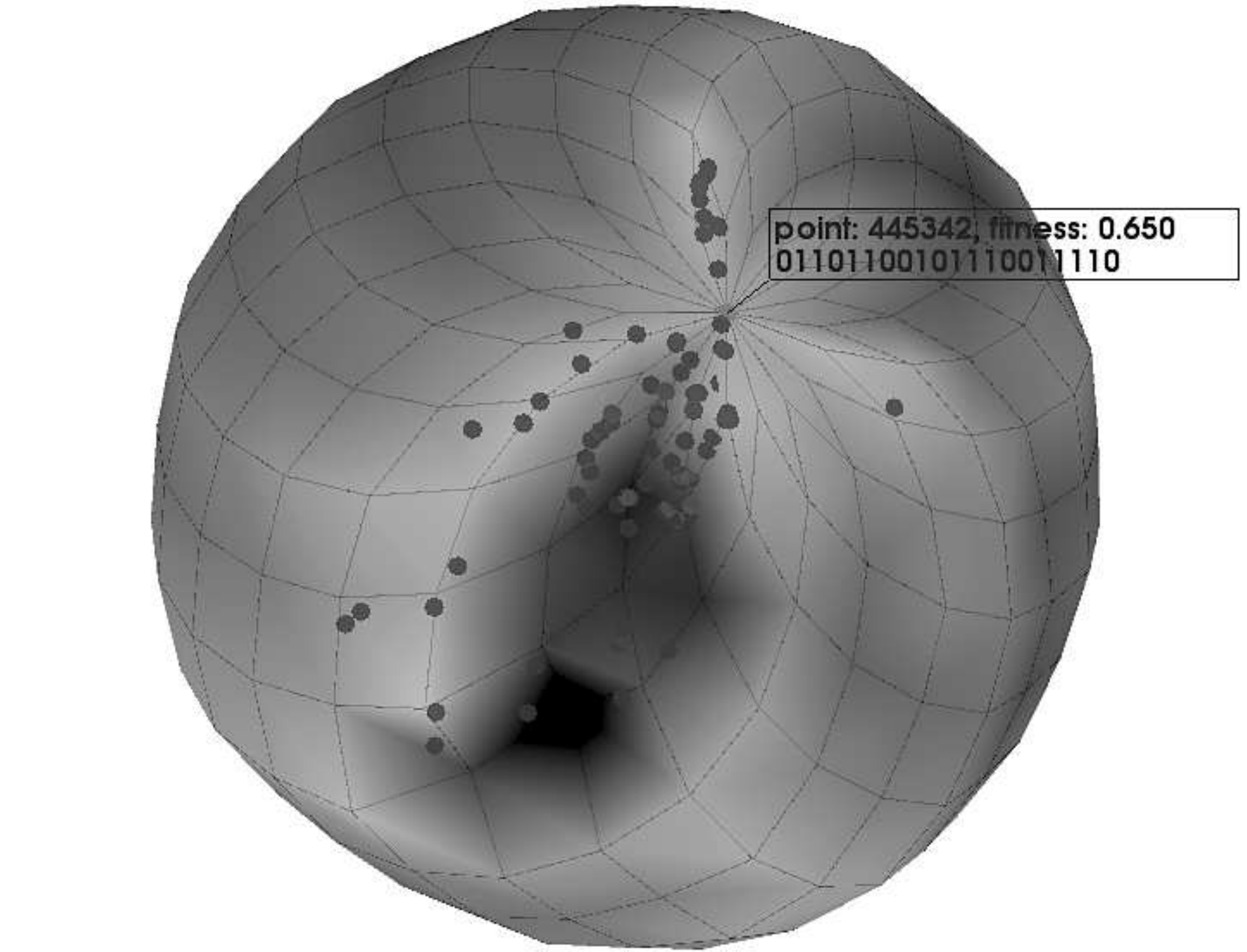}
\caption[]{A single agent's wave-riding behavior shown in the spherical visualization, with agent's past visited points marked with dark dots and its last position pointed to by the box (see Appendix~A for details of the visualization).
The agent follows a movement pattern of jumping to a point that has been partially boosted, then as the current point starts getting crowded, jumping to a similar nearby point. This behavior is clearest in sparse environments, demonstrating how agents can compete well by being constantly on the move. It is also similar to incremental improvements in many high-technology industries, giving them a computational interpretation. An animation of wave-riding behavior can be seen at \url{http://nn.cs.utexas.edu/?waveriding}.
}
\label{fig:waveriding}
\end{figure}

%> Wave-riding?
A particularly interesting behavior was observed in strategies evolved in sparse environments.
The probability of an agent encountering other agents in sparse fitness landscapes is rather small. Therefore, short exploitation jumps allow the agent to \emph{ride a boosting wave}, i.e.\ to stay at the forefront of the area that is being boosted as it moves through the landscape, leaving a trail of past visited points that have sunk in fitness (Figure~\ref{fig:waveriding}).
At each step, the agent boosts all nearby points in the neighborhood defined by the flocking radius. Therefore, as the new point starts losing fitness due to crowding, the agent can find a nearby point with similar fitness that is partially boosted, similarly to a surfer riding a wave.

Such \emph{wave-riding} behavior is particularly interesting because it resembles what happens in real-world innovation search. Companies often make slight improvements to their existing products, exploiting their initial design for many years without a significant redesign. The evolutionary simulations in this article rediscovered this strategy and demonstrated computationally why it is effective.

%%%%%%%%%%%%%%%%%%%%%%%%%%%%%%%%%%%%%%
%%%%%%%%%%%%%%%%%%%%%%%%%%%%%%%%%%%%%%

\section{Discussion and Future Work}
\label{sec:discussion}

The experiments in Section~\ref{sec:expEvolveParticular} showed that distinctly different search strategies evolve in different environments, and they are generally better than strategies evolved in other environments, evolved general strategies, and manual strategies. They are also significantly more complex than the manual strategies, and would be difficult to design without an automatic machine discovery method such as evolutionary computation. The main hypothesis of the article was therefore verified with these results: evolutionary computation is useful in discovering good strategies for CMAS problems.

The main areas of future work include developing more versatile strategy representations, applying the framework to analyzing real-world archival data, characterizing CMAS theoretically, and extending the framework with opponent modeling and communication.

First, the strategies were only coarsely encoded in this study. For instance, agents currently choose the best point of either the public memory or the private memory. A new action could be added to let the agents pick the best of both memories, depending on their fitness. The observation time of memory points could also be taken into account to allow the agents treat outdated points in memory differently. Furthermore, the power of CPPN to represent strategies was not yet fully utilized. Because their inputs and outputs are continuous, CPPNs could in principle represent strategies over very large number of (even continuous) states and actions. Such an approach would make it possible to represent much more refined strategies, which could in turn lead to more complex emergent behaviors, and to more accurate modeling of real-world search.

%> Homogen. in all + Heterogen.
In addition to evolving environment-specific strategies, Section~\ref{sec:heterogeneousEnv} showed how a single general strategy can be evolved to perform well across multiple different environments, with only a small cost in performance.
For such a strategy to scale up to even more diverse environments, it might be beneficial to allow multimodal behaviors, so that the same agent strategy can have distinct behaviors in different contexts \cite{schrum:cig11}.
A further step in this direction would be to coevolve all or a subset of the opponents as well. In this manner, the environment could present more diverse challenges, and more interesting and perhaps realistic general strategies could evolve.

More generally, the simulations in this article suggest that the CMAS approach could be used to provide insight into what kinds of strategies work well in real-world competitive multi-agent search. One possibility is to set up the search space and the agent parameters based on real-world archival data, such as the historical record on patents and products in a particular industry. Search can then be simulated on that landscape, explaining why certain strategies were effective, and potentially discovering new strategies that would have worked even better.

On the other hand, CMAS is a general and formally defined problem domain, which should make it possible to analyze it theoretically.
For example, stochastic processes could be used to characterize the scope and power of the search methods, deriving convergence and dominance conditions, as has been done in prior work on coevolution and estimation of distribution algorithms \cite{stanley:jair04,alden:phd07,ficici:ecal01,muhlenbein:ec05}.

Aspects of real-world competitive multi-agent search that were not addressed in this study include opponent modeling and explicit interactions between agents via direct communication. The former would let agents adapt to the environment by altering their strategies depending on the behavior of opponents they observe \cite{Carmel:1996}. The latter would allow agents to cooperate more effectively, form coalitions, and negotiate. These extensions would all be useful in modeling real-world innovation search, and constitute a most interesting direction of future work.

%%%%%%%%%%%%%%%%%%%%%%%%%%%%%%%%%%%%%%
%%%%%%%%%%%%%%%%%%%%%%%%%%%%%%%%%%%%%%

\section{Conclusion}
\label{sec:conclusion}

In this article, competitive multi-agent search was introduced as a formalization of human problem solving, with innovation search in organizations as a specific motivating example.  In this formalization, the agents interact through knowledge of other agents' searches and through the dynamic changes in the search landscape that result from these searches. The main contribution is to show that evolutionary computation is a useful method for CMAS problems: it is possible to discover effective search strategies that might be hard to design by hand, and understand why they are effective. CMAS thus demonstrates an interesting role for evolutionary computation: not only it can be used as an automated method for engineering, but also a way to understand how human behavior can be more effective. In the future, it may be possible to use CMAS simulations to make recommendations to human decision makers, as well as inform policy makers that aim at encouraging innovation and creativity.

%%%%%%%%%%%%%%%%%%%%%%%%%%%%%%%%%%%%%%
%%%%%%%%%%%%%%%%%%%%%%%%%%%%%%%%%%%%%%

%\section*{Acknowledgements}

%This research was supported in part by NSF under grants SBE-0914796, IIS-0915038, and DBI-0939454.

%%%%%%%%%%%%%%%%%%%%%%%%%%%%%%%%%%%%%%
%%%%%%%%%%%%%%%%%%%%%%%%%%%%%%%%%%%%%%

\appendix
\section*{Appendix A. Spherical \emph{NK} Fitness Landscape Visualization}

Although high-dimensional $N\!K$ landscapes are useful in testing ideas about search and optimization in complex domains, it is not possible to visualize them accurately, and it is therefore difficult to develop an intuitive understanding of what happens in such spaces. In a typical visualization, two dimensions are chosen to be represented accurately along each axis, and the visualization is repeated for the different combinations of values for the other dimensions. The main problem with such a visualization is that the continuity of the space is lost, i.e. nearby points can end up very far apart on the visualization, disallowing natural intuitions about space. Another issue with this kind of visualization is that it shows all $2^N$ points in the space; as $N$ grows, this number becomes prohibitively large, making it impossible to visualize $N\!K$ landscapes with sufficiently large $N$.

A different, novel approach is taken in Figures~\ref{fig:RTAConsideredPoints} and \ref{fig:waveriding}. In this spherical approach, the continuity of the space is preserved, and the space is represented with variable resolution. One point is chosen as the focus (e.g.\ 11111 in the five-dimensional case), and all of its neighbors in the original space (e.g.\ 01111, 10111, 11011, 11101, and 11110) are shown as its neighbors on the sphere, around a circle at distance 1. At distance 2, points with two bits away from the original are shown, by combining bit flips of the adjacent neighbors (to 00111, 10011, 11001, 11100, 01110), and so on until the complement (00000) of the focus point is reached at the other side of the sphere. In this manner, continuity of the original space is maintained in the grid that results on the sphere: nearby points in the grid are indeed neighbors in the original space, differing by one bit. The elevation and brightness of the spherical surface represents the fitness of the points on it: The higher the point and the lighter it is, the higher the fitness.

The continuity, however, imposes a trade-off: The grid is a complete representation of the space only near the poles (i.e. at distance 1 and 4 in the five-dimensional case); at the equator, it can only represent sample points in the space (i.e. at distances 2 and 3, only five of the ten points are located on the grid). For visualization (as in Figures~\ref{fig:RTAConsideredPoints} and \ref{fig:waveriding}), other points of interest can be shown in the quadrilateral regions between the actual grid points, although their specific locations within those regions are undetermined. They are placed so that the distance between each point and the north pole point corresponds to the Hamming distance between them. Among the quadrilateral regions along the meridian at that distance, each point is in the one whose corners are closest to the point.

The main purpose of this visualization is to represent the local neighborhood of a specific point intuitively as a continuous space, with gradually less resolution towards the horizon. It thus gives a concrete snapshot of the current state of the search. The focus point can also be moved as the search progresses, resulting in a detailed track of the process. Such a visualization tool is implemented in a software package \emph{NKVis}, i.e. a visualization tool for $N\!K$ fitness landscapes, which is freely available at \url{http://nn.cs.utexas.edu/?nkvis}.

%\section*{Appendix B. Simulation and NEAT Parameters}

\begin{table}[ht]
\begin{center}
{\normalsize\sc Appendix B. Simulation and NEAT Parameters}\\[2ex]
\rowcolors{1}{}{mylightgray}
\begin{tabular}{|l|l|}
\hline
\textbf{Simulation Parameter} & \textbf{Value} \\
\hline
$N$ (for the $N\!K$ model) & 10 and 20 \\
$K$ (for the $N\!K$ model) & 3 \\
Number of agents & 8 \\
Number of time steps per run & 100 \\
Number of runs per strategy evaluation & 200 \\
Flocking intensity & 1.05 $\rightarrow$ 0.9  \\
Flocking radius & 2  \\
Evolutionary generations & 500 \\
Evolutionary repeats per setup & 64 \\
Evolutionary population size & 100 \\
\hline
\rowcolor{white}
\multicolumn{1}{l}{} & \multicolumn{1}{l}{} \\
\hline

\textbf{NEAT Parameter} & \textbf{Value} \\
\hline
AddBiasToHiddenNodes & 1 \\
AdultLinkAge & 2 \\
AgeSignificance & 1.2 \\
AllowAddNodeToRecurrentConnection & 0 \\
AllowRecurrentConnections & 0 \\
AllowSelfRecurrentConnections & 0 \\
CompatibilityModifier & 0 \\
CompatibilityThreshold & 20 \\
DisjointCoefficient & 1.0 \\
DropoffAge & 10 \\
ExcessCoefficient & 1.0 \\
ExtraActivationFunctions & 1 \\
ExtraActivationUpdates & 19 \\
FitnessCoefficient & 1.0 \\
ForceCopyGenerationChampion & 1 \\
LinkGeneMinimumWeightForPhentoype & 0 \\
MutateAddLinkProbability & 0.2 \\
MutateAddNodeProbability & 0.2 \\
MutateDemolishLinkProbability & 0.04 \\
MutateLinkProbability & 0.2 \\
MutateLinkWeightsProbability & 0.8 \\
MutateNodeProbability & 0.05 \\
MutateOnlyProbability & 0.5 \\
MutateSpeciesChampionProbability & 0 \\
MutationPower & 2 \\
OnlyGaussianHiddenNodes & 0 \\
SignedActivation & 0 \\
SmallestSpeciesSizeWithElitism & 1 \\
SpeciesSizeTarget & 0 \\
SurvivalThreshold & 0.2 \\
WeightDifferenceCoefficient & 0.8 \\

%\hline

\end{tabular}
\end{center}
\caption{Parameters used in the experiments for the multi-agent simulation and NEAT.}
\label{tbl:parameters}
\end{table}%

%\end{appendices}

%%%%%%%%%%%%%%%%%%%%%%%%%%%%%%%%%%%%%%
%%%%%%%%%%%%%%%%%%%%%%%%%%%%%%%%%%%%%%

\bibliographystyle{IEEEtran}
\bibliography{nnstrings,nn,refs-rkatila,thispaper}

\end{document}